# Application of Machine Learning Methods in Inferring Surface Water – Groundwater Exchanges using High Temporal Resolution Temperature Measurements

**Mohammad A. Moghaddam[1], Ty P. A. Ferre[1], Xingyuan Chen[2], Kewei Chen[2], and Mohammad Reza Ehsani[1]**

[1] Hydrology and Atmospheric Sciences, University of Arizona, Tucson, AZ, USA - 85721

[2] Pacific Northwest National Laboratory, Richland, WA, USA

Emails: moghaddam@email.arizona.edu, tyferre@arizona.edu, xingyuan.chen@pnnl.gov, kewei.chen@pnnl.gov, rehsani@email.arizona.edu

Corresponding author: Mohammad A Moghaddam (moghaddam@email.arizona.edu)

# Abstract

We examine the ability of machine learning (ML) and deep learning (DL) algorithms to infer surface/ground exchange flux based on subsurface temperature observations. The observations and fluxes are produced from a high-resolution numerical model representing conditions in the Columbia River near the Department of Energy Hanford site located in southeastern Washington State. Random measurement error, of varying magnitude, is added to the synthetic temperature observations. The results indicate that both ML and DL methods can be used to infer the surface/ground exchange flux. DL methods, especially convolutional neural networks, outperform the ML methods when used to interpret noisy temperature data with a smoothing filter applied. However, the ML methods also performed well and they are can better identify a reduced number of important observations, which could be useful for measurement network optimization. Surprisingly, the ML and DL methods better inferred upward flux than downward flux. This is in direct contrast to previous findings using numerical models to infer flux from temperature observations and it may suggest that combined use of ML or DL inference with numerical inference could improve flux estimation beneath river systems.

**Keywords:** deep learning; machine learning; surface/ground water flux

## Highlights

- The ability of machine learning (ML) and deep learning (DL) algorithms to infer surface/ground exchange flux based on subsurface temperature observations is investigated
- DL methods, especially convolutional neural networks (CNN), outperform the ML methods when used to interpret noisy temperature data
- ML methods better identify a reduced number of important observations, which can be useful for measurement network optimization
- Unlike conventional methods, ML and DL better inferred upward flux than downward flux

# 1- Introduction

Estimating the exchange of water between surface and groundwater remains a critical and challenging task in groundwater hydrology. This interaction impacts subsurface water availability and quality, the health of river systems, and the interaction between surface and subsurface water stores. Several studies (Pionke et al.; Triska et al.; Sophocleous) have shown that this exchange exerts strong controls on stream chemistry, including dissolved oxygen levels, nutrient levels, pH, and other water quality measures. Direct measurement of water flux remains challenging: seepage meters (Rosenberry et al.) can provide flux measurements, but they require considerable effort, provide limited temporal resolution, and are not readily automated; point velocity probes (Labaky et al.) show great promise for high-resolution velocity measurement but are limited in temporal resolution in highly transient environments. As a result, the surface exchange flux must be inferred indirectly. Temperature-based methods have been developed to infer surface/ground water exchange. Some methods, such as heat pulse probes (Lewandowski et al.), use active sources, generally providing accurate measurements, but requiring relatively complex, multi-rod instruments for each observation location. Methods that rely on natural temperature variations to infer water flux through its impact on advective heat transport have found wider use because of their relatively simple measurement support needs – often only requiring temperature time series at several depths (Constantz; Essaid et al.; Vandersteen et al.; Kikuchi and Ferré; Munz and Schmidt). This approach is supported by data collection using sensors ranging from simple thermistors to fiber-optic cables (Mamer and Lowry; Hare et al.). To date, interpretation of temperature records for water flux estimation has focused on the use of analytical or numerical models that describe coupled fluid flow and heat transport (Healy and Ronan; Therrien et al.; Becker et al.; Hatch et al.; Keery et al.; Anibas et al.; Briggs et al.; Cuthbert and MacKay; Mamer and Lowry; Onderka et al.; Voytek et al.). Numerical approaches have greatly advanced our ability to infer flow, but calibration can require supporting information such as sediment thermal parameter values, and flux estimates are generally limited to providing daily average fluxes. Further restrictions apply when using analytical solutions, including homogeneity, 1D vertical flux, and slowly varying fluxes (López-Acosta). A more complete discussion of the limitations of the analytical approach is provided by (Rau et al.).

All methods that aim to infer flux based on natural temperature variation assume that the model used to interpret the temperature data can recover enough relevant information from the temperature observations to infer the flux. Recognition of the limitations imposed by the assumed model led from the initial use of analytical models to later use of numerical models. Recently, there has been an explosion of interest across scientific and engineering disciplines in the use of data-driven models, specifically machine learning methods, to provide a model-free approach to data analysis (Ehsani, Arevalo, et al.; Arabzadeh et al.; Gupta et al.; Ehsani and Behrangi; Song et al.; Aldhaif et al.; Angelaki et al.; Coley et al.; Rohmat et al.). ML methods uncover relationships among state variables without explicit knowledge of the system (Adhikari et al.). Given sufficiently informative data, they can mitigate some limitations of physics-based models, including imperfect knowledge of system structure and parameter values (Ehsani, Arevalo, et al.). In hydrogeology, several attempts have been made to learn governing equations and partial differential equations from input and output data (Triana et al.; Chang and Zhang; Rohmat et al.). In particular, (Chang and Zhang) investigated the possibility of using the least absolute shrinkage and selection operator (LASSO) to learn the 1-D subsurface flow equation using synthetic data generated by a numerical, physics-based model. (Triana et al.) trained a single layer feedforward network as a surrogate for a MODFLOW model to characterize a stream-aquifer system affected by irrigation. Most directly applicable to the current study, (M. Moghaddam et al.) examined the potential uses of simple machine learning algorithms (decision tree and gradient boosting) to infer surface/ground water flux based on synthetic data produced by a numerical flow and heat transport model. That study examined the use of simple, tree-based machine learning (ML) models to extract water flux from temperature measurements with high temporal resolution (5 minutes) in a highly dynamic system. This study extends that work to investigate the use of more advanced ML techniques to infer surface exchange from subsurface temperature observations.

There are many machine learning algorithms, ranging from very simple clustering techniques (Tang et al.; Xu et al.; Javadi et al.; Dadashazar et al.) to more complex neural networks (Ehsani, Zarei, et al.; M. A. Moghaddam et al.; Bouktif et al.; Kratzert et al.; Afzaal et al.). In some cases, simple algorithms are preferred because they are rapid, simple to implement, and highly informative regarding the value of observations (Ehsani, Behrangi, et al.). This latter quality arises from measures of the value of specific observations to the predictions that are inherent to the tree-based algorithms and may be particularly useful for investigations that include an objective of optimizing the observation network. But, simple ML tools can be limited in application to nonlinear problems (Adhikari et al.; Ehsani, Arevalo, et al.; Ehsani, Behrangi, et al.; Vezhnevets and Barinova; Georganos et al.). Deep learning (DL) is a subset of machine learning algorithms that learns representations of the data with multiple levels of abstraction. Due to recent advances in storage and computational power and the continued improvement of DL algorithms, these methods are rapidly becoming the standard for scientific and engineering applications (Ehsani, Zarei, et al.; Najafabadi et al.; Aliper et al.; Shen). More recently, DL has found applications in hydrology and water resources management (Ehsani, Zarei, et al.; Ehsani, Behrangi, et al.; Adhikari et al.; Ehsani, Arevalo, et al.; M. A. Moghaddam et al.; Kratzert et al.; Afzaal et al.; Assem et al.; Pelissier et al.). The specific application of deep learning to surface/ground water exchange is limited. In some recent examples, (Rohmat et al.) used a multi-layer perceptron (MLP) model as a computationally-efficient surrogate for a MODFLOW-SFR2 model to assess basin scale impacts of best management practice implementations on stream-aquifer exchange and water rights. (Afzaal et al.) estimated groundwater level fluctuations using DL algorithms with stream level inputs.

The power of DL models over other classic ML methods lies in their ability to learn more complex functions while providing enhanced generalization capabilities (Ehsani, Behrangi, et al.; Ehsani, Zarei, et al.). But, despite their predictive power, these models can suffer from a lack of interpretability (Apley and Zhu; Chakraborty et al.). The present study has three objectives that relate to the capabilities and limitations of DL algorithms. First, we examine the potential for DL algorithms to infer surface/ground water flux from subsurface temperature observations with high temporal resolution (5 min) in a highly dynamic system; we compare the performance of these more advanced ML methods to that of the simple tree-based methods examined by (M. Moghaddam et al.). Second, we examine the impact of measurement noise on the performance of these DL tools. In particular, we test whether temporal filtering techniques can be combined with more advanced ML/DL algorithms to minimize the impacts of temporally uncorrelated measurement noise. Finally, we examine approaches to extract information from DL methods that can be used to design efficient and effective monitoring networks.

# 2- Methodology

This investigation represents an initial feasibility study of deep learning methods for interpreting surface/ground water exchange from subsurface temperature measurements collected with high temporal resolution. A highly resolved numerical model of water flow and heat transport was driven by dynamic river stage and surface water temperature to produce time series of surface/ground water exchange flux and temperature at multiple depths. The forecast targets for the machine learning algorithms were the flux time series; the subsurface temperatures at multiple depths were used to develop features (i.e., model inputs). The input features (derived from the temperature observations) were: temperature; time delayed temperature; and temperature gradients in space and time.

We considered multiple DL algorithms: Multilayer Perceptron (MLP), Long Short Term Memory (LSTM), and Convolutional Neural Network (CNN). The performance of each method was compared to that of XGBOOST, a prominent tree-based machine learning algorithm that has shown strong performance in regression competitions (Chen and He; de Vito). XGBOOST is an optimized implementation of gradient boosting that penalizes complexity to reduce overfitting in the case of noisy data (Chen and He; Chen and Guestrin). We expected the deep learning algorithms to show better performance; we included the classical method as a baseline for evaluating performance improvement using deep learning. XGBOOST has an intrinsic measure of feature importance that can be used to design an optimal reduced observation set (M. Moghaddam et al.). An external feature importance estimation method, Accumulated Local Effects analysis (Apley and Zhu), was applied to both XGBOOST and the DL algorithms in an attempt to identify the observations that were most informative for inferring the surface/ground water exchange flux. Analyses were performed with and without added measurement noise. In addition, several denoising filters were used to improve the performance of the ML tools when interpreting data with added noise.

## 2-1- Numerical Model

In this study, the synthetic observation data were generated from a 1-D column model using boundary conditions extracted from a 3-D hydrothermal model developed at the Department of Energy Hanford site located in southeastern Washington State, (Figure 1A). The coupled flow and heat transport equations were solved by PFLOTRAN, which is a massively parallel multiphase flow and reactive transport simulator implemented in object-oriented FORTRAN (Hammond et al.; Lichtner et al.). The underlying equations representing flow and heat transport are:

$$\frac{\partial}{\partial t}(\varphi s\eta) + \nabla \cdot (\eta q) = Q_w \quad [1] \qquad \textit{(Equation-1)}$$

$$\frac{\partial}{\partial t}(\varphi s\eta U + (1-\varphi)\rho_r c_p T) + \nabla \cdot (\eta q H - \kappa \nabla T) = Q_e [2] \qquad \textit{(Equation-2)}$$

In Eq. [1], $\boldsymbol{\varphi}$ is the porosity (dimensionless), $\boldsymbol{s}$ is the saturation (dimensionless), $\boldsymbol{\eta}$ is the molar water density ($NL^{-3}$), $\boldsymbol{q}$ is the Darcy flux ($LT^{-1}$), and $\boldsymbol{Q_w}$ represents all sources and sinks ($MT^{-1}$). In Eq. [2], $\boldsymbol{U}$ and $\boldsymbol{H}$ are internal energy ($J$) and enthalpy of fluid ($J$), $\boldsymbol{\rho_r}$ is rock density [$ML^{-3}$], $\boldsymbol{c_p}$ and $\boldsymbol{\kappa}$ are the heat capacity ($J\Theta^{-1}$) and thermal conductivity ($MLT^{-3}\Theta^{-1}$) of water, and $\boldsymbol{Q_e}$ is the heat source($JT^{-1}$). The viscosity as a function of temperature and pressure can be defined by Eq. [3].

$$\mu_w = 241.4 * 10^{247.8/(\mathrm{T}-140)}(1.0 + 1.0467 \times 10^{-6} (\mathrm{p} - \mathrm{p_{sat}})(\mathrm{T} - 305)) \qquad \textit{(Equation-3)}$$

Where $\boldsymbol{\mu_w}$ is water viscosity in μP; $\boldsymbol{T}$ is the temperature in K; $\boldsymbol{p}$ is pressure in bars and $\boldsymbol{p_{sat}}$ is saturation pressure in bars corresponding to temperature $\boldsymbol{T}$(Meyer).

A schematic 3D representation of the model is shown in (Figure 1A). The model has dimensions of 400m × 400m × 20m with a refined mesh size of 0.5m × 2m × 0.1m near the river. Long-term observations of the temperature and stage of the river were used as the river boundary condition. Details for this larger scale model are available in (Bisht et al.). A refined 1D model, 2m in vertical extent, was developed based on the results of the 3D model. This high-resolution model had 0.01m cell resolution. The top boundary was defined by the time series of stage and temperature; the bottom of the domain was a Dirichlet boundary for flow and transport with water pressure and temperature extracted from the 3D model (Figure 1B). The 1D model generated surface/groundwater flux and subsurface temperature and pressure every 5 minutes for 1.5 years. Following the study of (Chen et al.), we considered three shallow observations depths to represent a plausible monitoring network with sensors at 0.005, 0, 0.15, and 0.255 m depth. Unlike analytical or numerical models, the ML tools do not impose bottom boundary conditions. Therefore, it was assumed that they need some information to describe the influence of flow and heat transport across the bottom boundary. To provide this, a single deep observation was selected at 1.995m depth (Figure 1B).

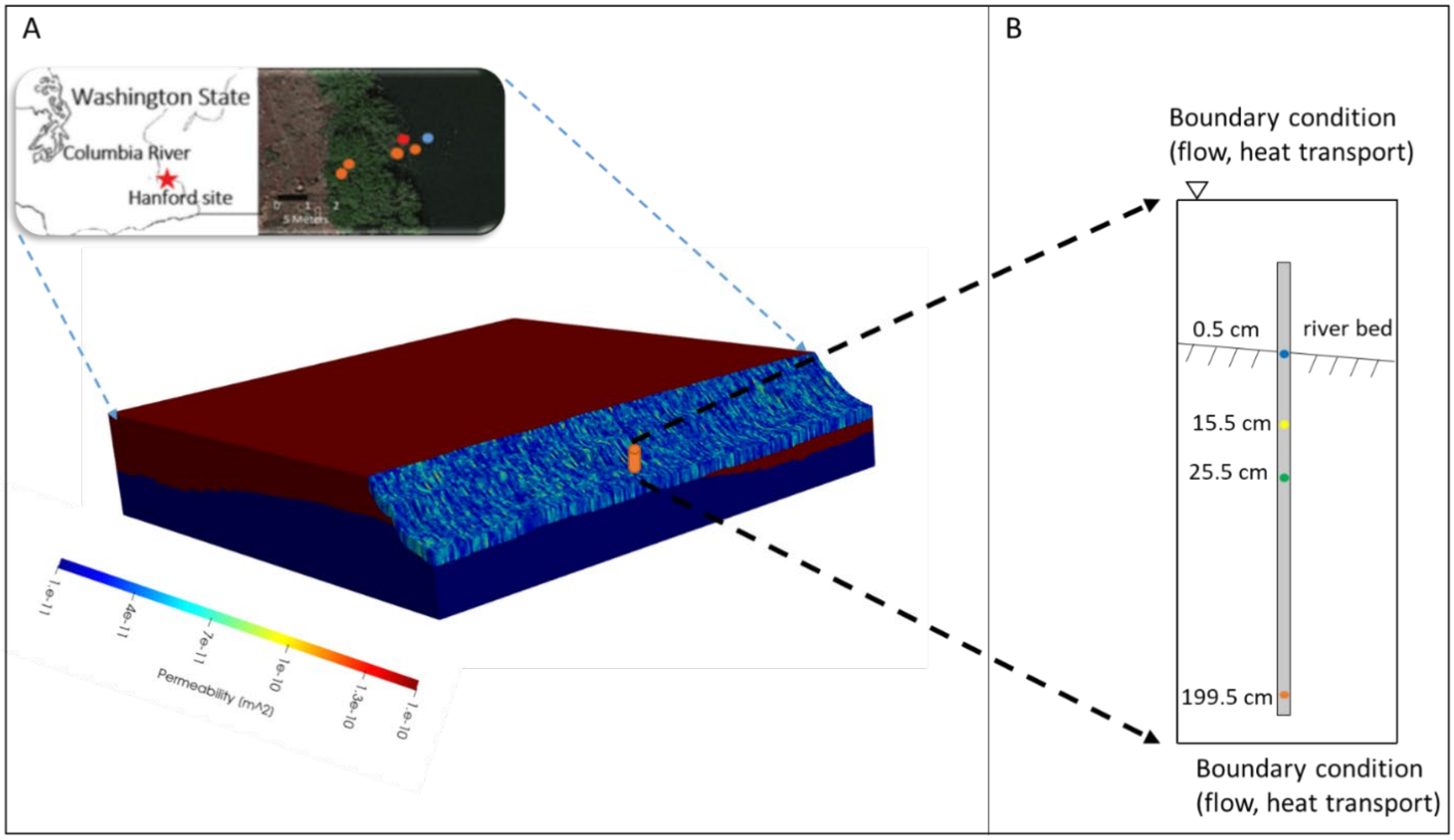

*Figure 1- A: 3D flow and heat transport model built for the study site, B: the 1D column model that mimics the Installed thermistor array. Color spot is the location where the thermistor array is used in ML analysis. The dimension for the 3D model is 400m x 400m x 20m in XYZ directions and the height of the column model is 1.995m corresponding to the length of the thermistor array.*

## 2-2- Data Preprocessing

Following a common practice for ML analyses, the data were divided into training, validation, and testing subsets. The validation set was used for hyperparameter tuning, while the testing set was used for the final assessment of the algorithms performance. (After tuning the models using the validation set, the models were retrained using the combined training and validation sets. With this arrangement, 70% of the data were used for training and 30% for testing.) When defining the testing and training sets, it is important to ensure that both the training and testing sets are representative of the underlying data generating processes. In this case, because the hydraulic conductivity depends on the temperature and the flux depends on the hydraulic conductivity and the pressure gradients, the training set had to be chosen to explore both the full range of surface fluxes and the full range of temperatures. Following the work of (M. Moghaddam et al.), we divided the 110,000 observation times into six alternating training-validation/testing periods (Figure 2). Training and validation were performed on observations: 500- 12500; 19000 - 25000; 33000 – 45000; 52000 – 70000; 75000- 90000; 97000- 110000 (shown in blue). Testing was performed on the remaining observations (shown in red). We used the latest-occurring 20% of each training window to form the validation set.

The only direct observations were the temperature time series, $\boldsymbol{obs_{temp,t}}$, at each sensor depth. But, we recognized that the infiltration flux at time $\boldsymbol{t}$ would not necessarily be reflected in the temperature at depth $z$ at time $t$. Some DL algorithms, such as LSTM, have been designed to accommodate time delays in information transfer throughout a system. Other methods require that these delays be introduced by manipulating the input to the ML/DL. To consider the time delay between surface flux and subsurface temperature, we provided the ML/DL tools temperature observations within a time window around the time of surface flux inference. Temperature observations collected after some time delay, are indicated with positive time subscripts (e.g. $t+3\Delta t$). For most surface water/ groundwater applications, it is expected that the data will be collected and then analyzed long (i.e., long after the data collection time window). As such, it is acceptable to use "*future*" observations for inference of fluxes that occurred before the time for which flux is being inferred.

To avoid unintentional bias, we also considered temperatures before the time of flux estimation, shown with negative subscripts. The delay between surface flux and subsurface temperature will depend on the depth of the sensor and the direction and magnitude of the water flux. Therefore, multiple time delays were considered simultaneously; later analysis attempted to determine which time delays were most informative for flux estimation. For these analyses, we considered temperature observations within a 1-hour window centered on the time of flux estimation. In choosing the window width, we balanced several considerations. The window had to be long enough to accommodate heat transport to the third sensor at 25.5 cm depth. However, the stated goal of the study is to identify flux with high temporal resolution; given the highly dynamic nature of the flux at the site (Figure 2A), it was considered to be unlikely that a very wide time window would be useful. Finally, it is critical that training and testing observations be isolated; as a result, it is not practical to consider very long time delays. Further discussion of the choice of the time window is presented in the Discussion. In addition to considering temperature time series, we used spatial, $\Delta \boldsymbol{d}_{\boldsymbol{temp,t}}$ and temporal, $\boldsymbol{\Delta t}_{\boldsymbol{temp,t}}$, gradients of the temperature as features. To reflect practically achievable gradient observations, spatial gradients were calculated between the sensors; that is, the addition of gradient measure would not require an additional subsurface sensor. We assumed that measurements were collected regularly, every 5 minutes, at any selected depth. The temporal gradient at a given depth was calculated between consecutive measurements in time. The input manipulations (feature engineering) are illustrated in Figure 3.

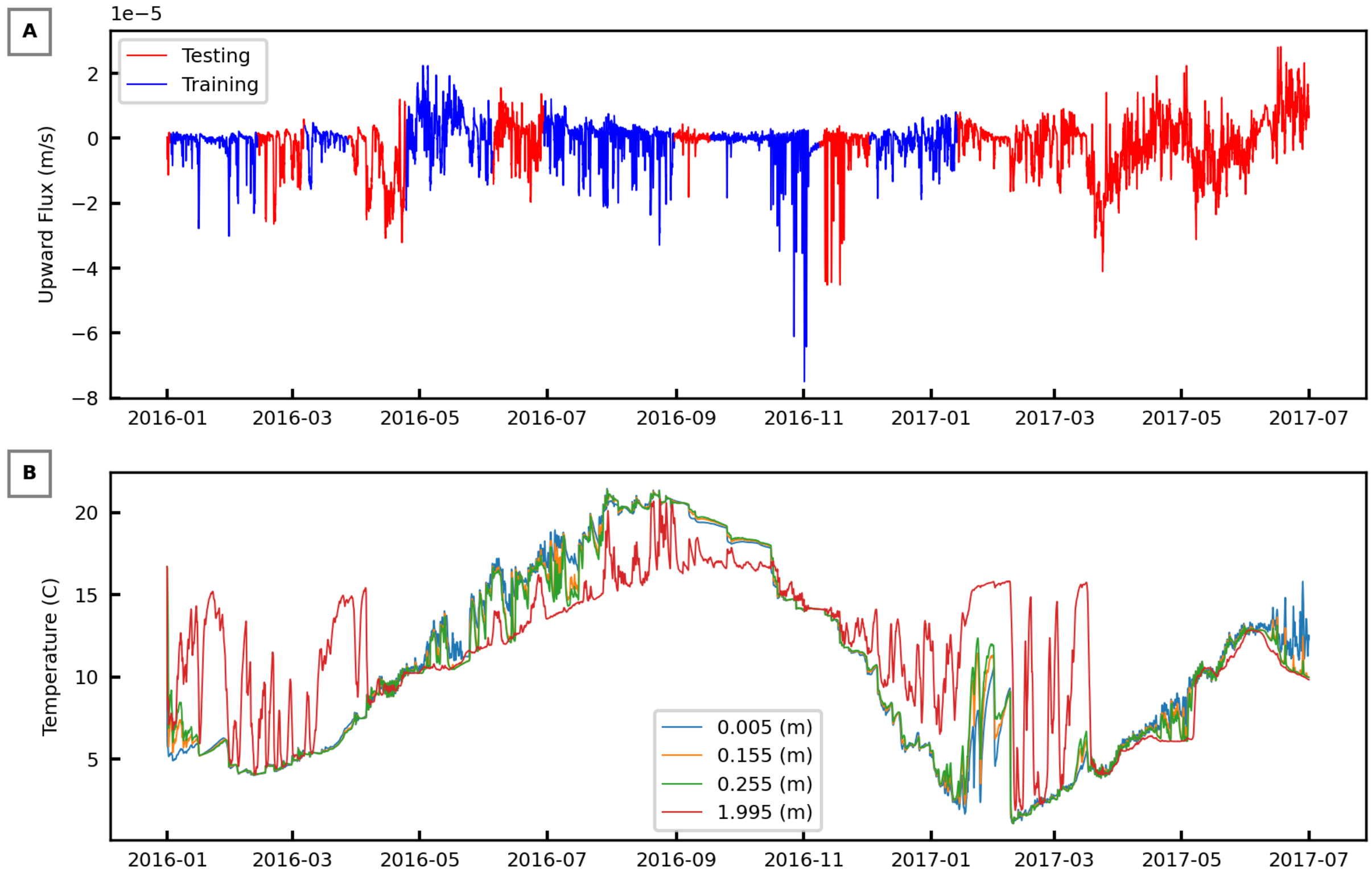

*Figure 2- Time series of observations A: Flux training (blue) and testing (red) sets illustrated using data from 0.005m depth. B: temperature at four depths*

ML/DL tools, like all inference methods, are affected by noise in the input data. In some cases, noise has been shown to improve the performance of ML/DL by reducing overfitting (Bishop; Noh et al.). More often, added noise reduces performance, with different impacts on each ML method(Nettleton et al.; Luo and Yang). An advantage of using synthetic data is that the level of noise can be controlled. We assume errors are caused by several independent sources in an additive manner, represented by a Gaussian noise model (Montgomery). Thus, we applied a generic error by superimposing zero mean Gaussian random errors with a standard deviation equal to a given percent of the variance of all noise-free temperature measurements at all depths and times. These errors are described as a signal to noise ratio (SNR), with SNR defined as the inverse of the ratio of the measurement error to the variance over all measurements of a given measurement type. For example, if the variance of the error-free observations is 100 times the variance of the added errors, then the SNR is reported as 100. A measurement set with an SNR of 10 would be considerably noisier than one with an SNR of 100.

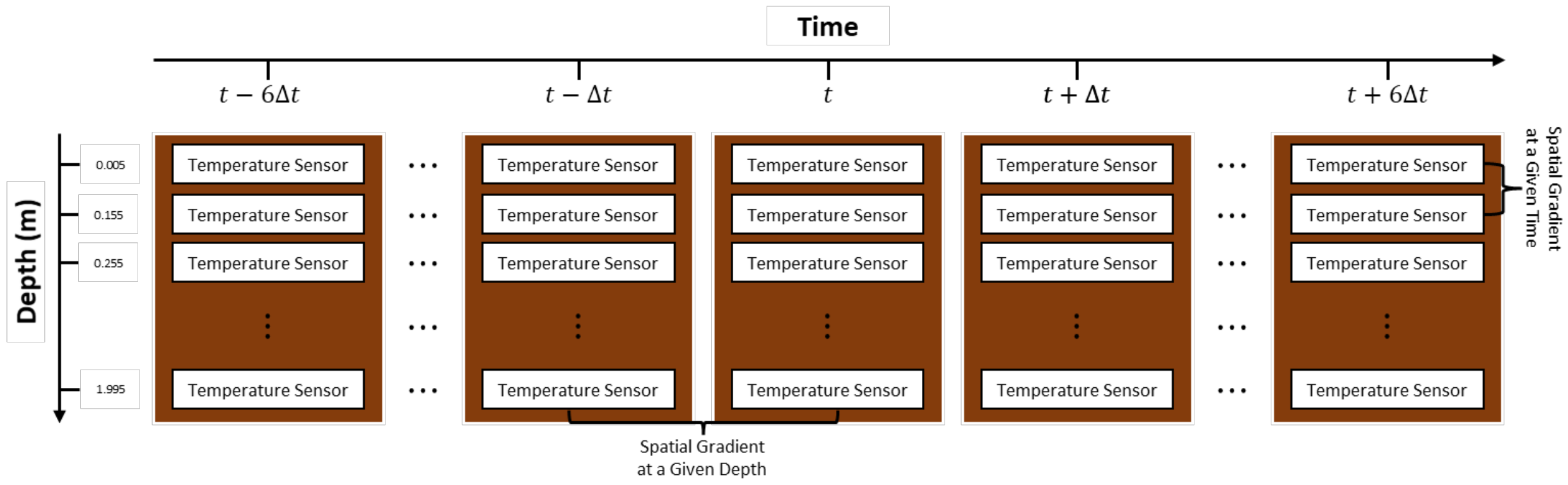

*Figure 3- Feature engineering schema for calculating spatial and temporal temperature gradients.*

## 2-3- Filtering and Smoothing

Smoothing is a technique to extract the main pattern from noisy data. This method is based on the convolution of a scaled window function with the signal. Window functions are symmetric about the middle of the interval with zero value outside of the interval. The performance of a technique depends on the window length, which should be tuned via trial and error. The window functions were rolled through the data by applying them to each overlapping segment of temperature time series with a step of one observation time (5 minutes). Eq. [4] describes the convolution operation for the noisy temperature data at time t for a window function with length N with $f'$representing a scaled window function.

$$(f * T)[t] = \sum_{n=-N/2}^{N/2} f'(n)T(t-n) \qquad \textit{(Equation-4)}$$

To have the same length of data before and after preprocessing while not allowing information to be shared by the training and testing pools, we used zero padding within one half of the window width of the beginning and end of each training or testing window. Training and testing were denoised separately to avoid blending of information at the edges of the training/testing windows. The window size was determined by grid search for each filter type, choosing the window size that produced the best performance for predicting the flux for the training sets.

After examining the performance of each ML/DL on data with added error, we filtered the inputs and repeated the analyses for every combination of filter type and ML/DL algorithm. In this study, we used Hamming, Flat (i.e., moving average), Blackman, Bartlett and Hanning filters. All of these are well known fixed-parameter window smoothing functions (Angrisani et al.; Zolfaghari et al.). The simplest, moving average, replaces each point with the average of adjacent data points within the window. Blackman, Hamming, and Hanning are cosine-shaped window functions that reduce high frequency signal oscillations induced by noise. In comparison to Hanning and Hamming, Blackman has a wider main lobe, which results in a higher pass band, but it has a faster attenuation rate. Hamming is more effective at canceling the nearest side lobe frequencies while Hanning is more desirable for noise cancelation for the rest of side lobe. Barlett is a linear, triangle shaped window function with zero values at the beginning and end of the function. The Bartlett window produces a monotonically decreasing frequency response magnitude in the frequency domain. In other words, the side lobe decreases faster for higher frequencies. Table [1] defines the equations for the window functions used in this study.

*Table 1- Window function equations.*

| **Filter Name** | **Equation** |
|---|---|
| Flat | $f(n) = \begin{cases} 1/N, & -N/2 < n < N/2 \\ 0, & Otherwie \end{cases}$ |
| Hanning | $f(n) = \begin{cases} 0.5 + 0.5\,.\cos\,(\frac{2\Pi n}{N})/, & -N/2 < n < N/2 \\ 0, & Otherwie \end{cases}$ |
| Hamming | $f(n) = \begin{cases} 0.54 + 0.46\,.\cos\,(\frac{2\Pi n}{N}), & -N/2 < n < N/2 \\ 0, & Otherwie \end{cases}$ |
| Blackman | $f(n) = \begin{cases} 0.42 - 0.5\,.\cos\left(\frac{2\Pi n}{N}\right) + 0.08\,.\cos\,(\frac{4\Pi n}{N}), & -N/2 < n < N/2 \\ 0, & Otherwie \end{cases}$ |
| Bartlett | $f(n) = \begin{cases} 1 + \frac{2n}{N}, & -\frac{N}{2} < n < 0 \\ 1 - \frac{2n}{N}, & 0 < n < N/2 \\ 0, & Otherwise \end{cases}$ |

## 2-4- Machine Learning Models

(M. Moghaddam et al.) compared the performance of simple tree-based machine learning techniques: decision tree and gradient boosting. In this section, we present a brief overview of the learning algorithms examined in this study and their associate hyper parameters.

### 2-4-A- Extreme Gradient Boosting (XGBOOST)

XGBOOST is a modified version of gradient boosting. Boosting is an ensemble technique to create strong regressors from a number of weak learners. In the boosting-based ML methods, models are built sequentially by weighting the error of prediction in all previous models to emphasize more difficult-to-predict targets, subsequently decreasing the overall error of the model. In gradient boosting, shallow decision trees are used as the weak learners to correct the prediction error of prior models. Unlike gradient boosting, XGBOOST uses a regularized model formalization to control overfitting. Also, XGBOOST has better performance and speed due to its distributed and parallel processing (Chen and He; Chen and Guestrin; Nielsen). A more detailed description of gradient boosting and XGBOOST can be found in (Natekin and Knoll; Chen and Guestrin; Chen and He).

### 2-4-B- Multilayer Perceptron Model (MLP)

MLP is a multilayer feed forward neural network that is considered to be the simplest form of a neural network. Each layer consists of several processing units (neurons). Each neuron is connected to adjacent layers with an individual weight assigned to each interlayer link. All inputs into a neuron are multiplied by their associated weight and summed to form a single output. Finally, each of these outputs is subject to a nonlinear transformation referred to as the activation function. As a result, MLP can be represented as a nested set of functions, $\boldsymbol{f_i}()$ . For example, a three-layer network can be defined as:

$$y = f_{NN}(x) = f_3(f_2(f_1(x))) \qquad \textit{(Equation-5)}$$

In which, $f_i()$ are vector functions of the following form:

$$f_1(z) \stackrel{\text{def}}{=} g_l(W_l z + b_l) \qquad \textit{(Equation-6)}$$

With $\boldsymbol{W_l}$ representing the weights and $\boldsymbol{b_l}$ the biases, and the activation function is $\boldsymbol{g_l}$. The architecture of an MLP, Figure 4, is commonly shown with the units within a layer connected to units in adjacent layers by arrows. The input features to the system are shown as circles (in our case, temperature times series, time delayed temperature time series, and temporal and spatial gradients of temperature with time).

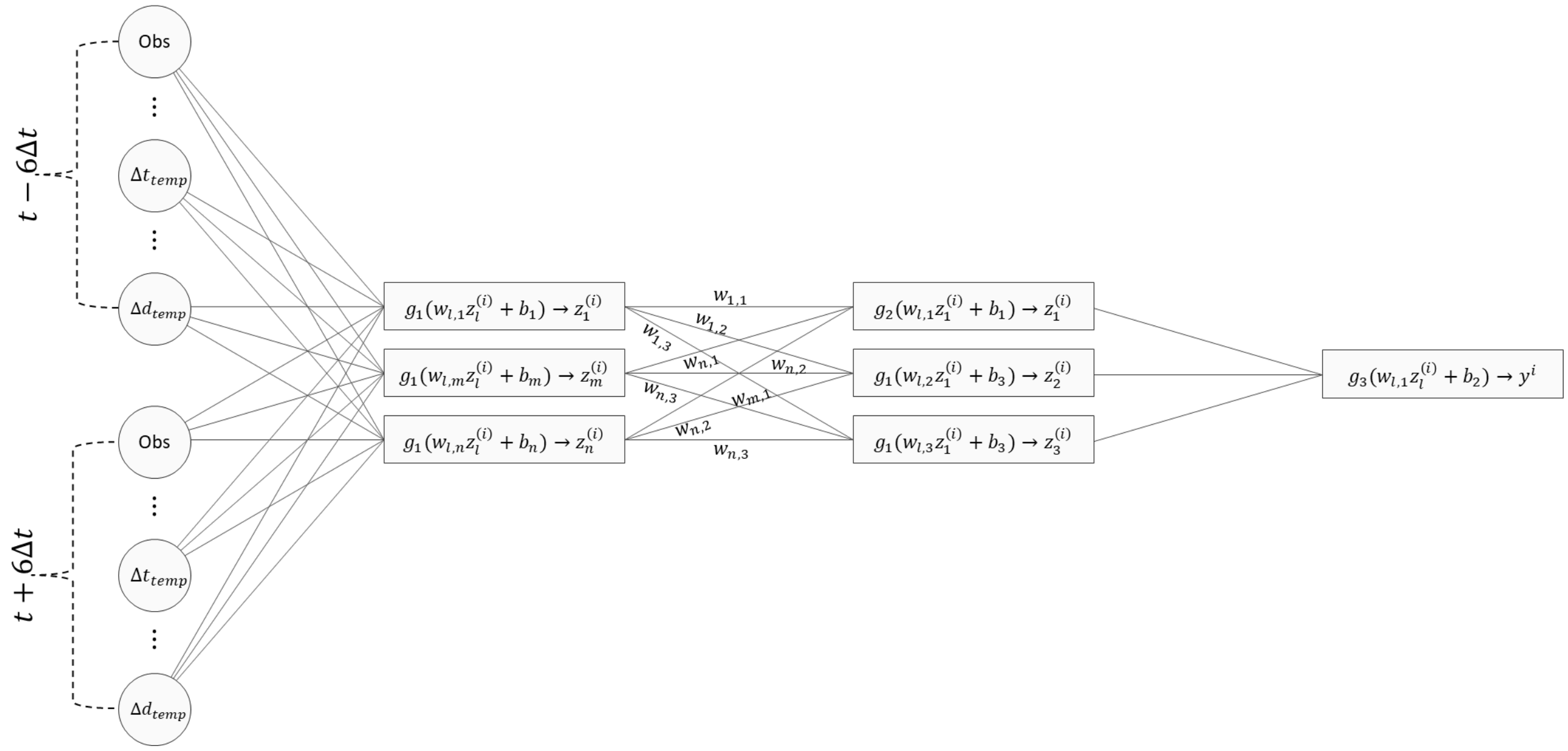

*Figure 4- Overall architecture of a multilayer perceptron (MLP) with weights of w, and two hidden layers with n hidden units in the first layer and 3 in the second layer. Model inputs are a 1-D vector consisting of designed features for t-Δ6t through t+Δ6t, and the outputs (y(i)) are a vector of streambed exchange fluxes, in which i represents the sample number for time t.*

### 2-4-C- Convolutional Neural Networks (CNN)

A CNN is a feedforward neural network that consists of two main components, a convolutional kernel and pooling layers. A kernel is a sliding window matrix or vector of weights, which is applied to a subset of the elements of the matrix of values being propagated through the network (the input matrix). The convolutional step is a dot product of the kernel and the input matrix. This acts like the weight functions in the MLP, but each entry in the matrix can now be influenced by neighboring values. As for the MLP, this step is followed by the application of an activation function. For the efficiency of training, the kernel weights are constant. The advantage of the CNN is that the convolution connects multiple elements of the matrix, but because the size of the kernel is generally small, there is sparse interaction between layers. Pooling is an optional subsampling operation that shrinks the size of a layer by replacing inputs with summary statistics. A CNN gains efficiency by sharing weights, applying sparse connectivity, and pooling. More on CNNs is available in (Selvin et al.; Zhao et al.).

Figure 5 shows the kernel used in our CNN. We used a 1D CNN, applied over all features at a single time step to avoid applying a convolution over both spatial and temporal features together. The 1 D kernel strides with step size one along the temporal direction. Temporal and spatial gradients were set to zero at the first observation time and shallowest observation depth. Trial and error tuning identified five layers, including the output layer, as optimal. The first two layers were convolutional layers followed by three fully connected layers, including the output layer (Figure 6). Because we did not observe significant overfitting, we did not use a pooling layer.

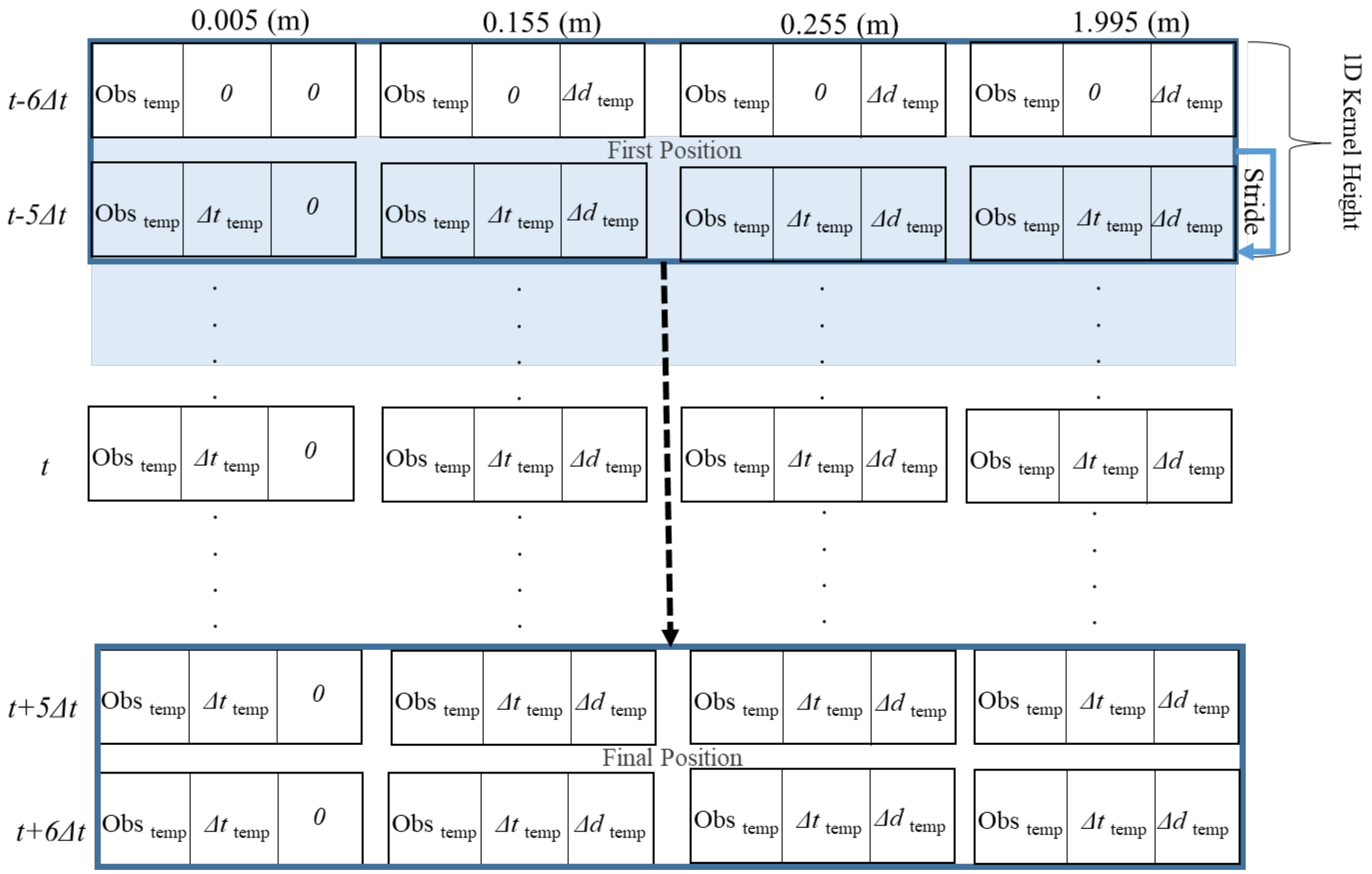

*Figure 5- General architecture of a 1D CNN. Model inputs are a 2-D array with time step as the rows and features as the columns, and the outputs (y(i)) are a vector of streambed exchange fluxes, in which i represents the sample number for time t.*

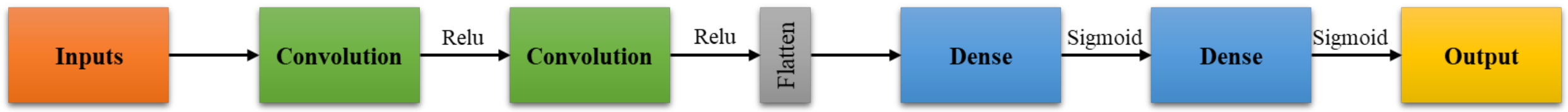

*Figure 6- The CNN architecture used in this study*

### 2-4-D- **Long Short-Term Memory Recurrent Neural Networks (LSTM)**

LSTM is a recurrent neural network. This architecture is particularly well-suited to represent the dynamic behavior of complicated sequences by considering the order of the inputs (e.g. a time series). By introducing memory and states cells into a recurrent neural network, an LSTM model can extract information from long-term dependencies in sequential data and minimize the vanishing gradient problem in longer sequences (Hochreiter and Urgen Schmidhuber). Figure 7 illustrates the structure of one LSTM cell. The information from previous (or later) time steps is transported by hidden and cell states. A hidden state is the output vector of an LSTM cell. A cell state directly conveys the information stored in long-term memory. The cell state is subject to different LSTM cells. Forget gates decide what information should be discarded from the cell state based on inputs and previous hidden states. Input gates decide what information should be updated according to the current inputs of an LSTM cell. Specifically, the current input and hidden states are passed through a sigmoid activation function to determine which entries of the cell state should be updated. Then, those inputs are passed to a *tanh* function to calculate the amount by which they should be updated. Finally, the output gate calculates the updated value of the next cell hidden state by applying a *sigmoid* function to the previous cell's hidden state and the current inputs multiplied by *tanh* of the updated cell state. In this study, we used a Bidirectional LSTM (BiLSTM), which feeds the information forward in time first and then backward in time. The hidden states of the forward and backward sequences are averaged and then fed into a fully connected (FC) layer.

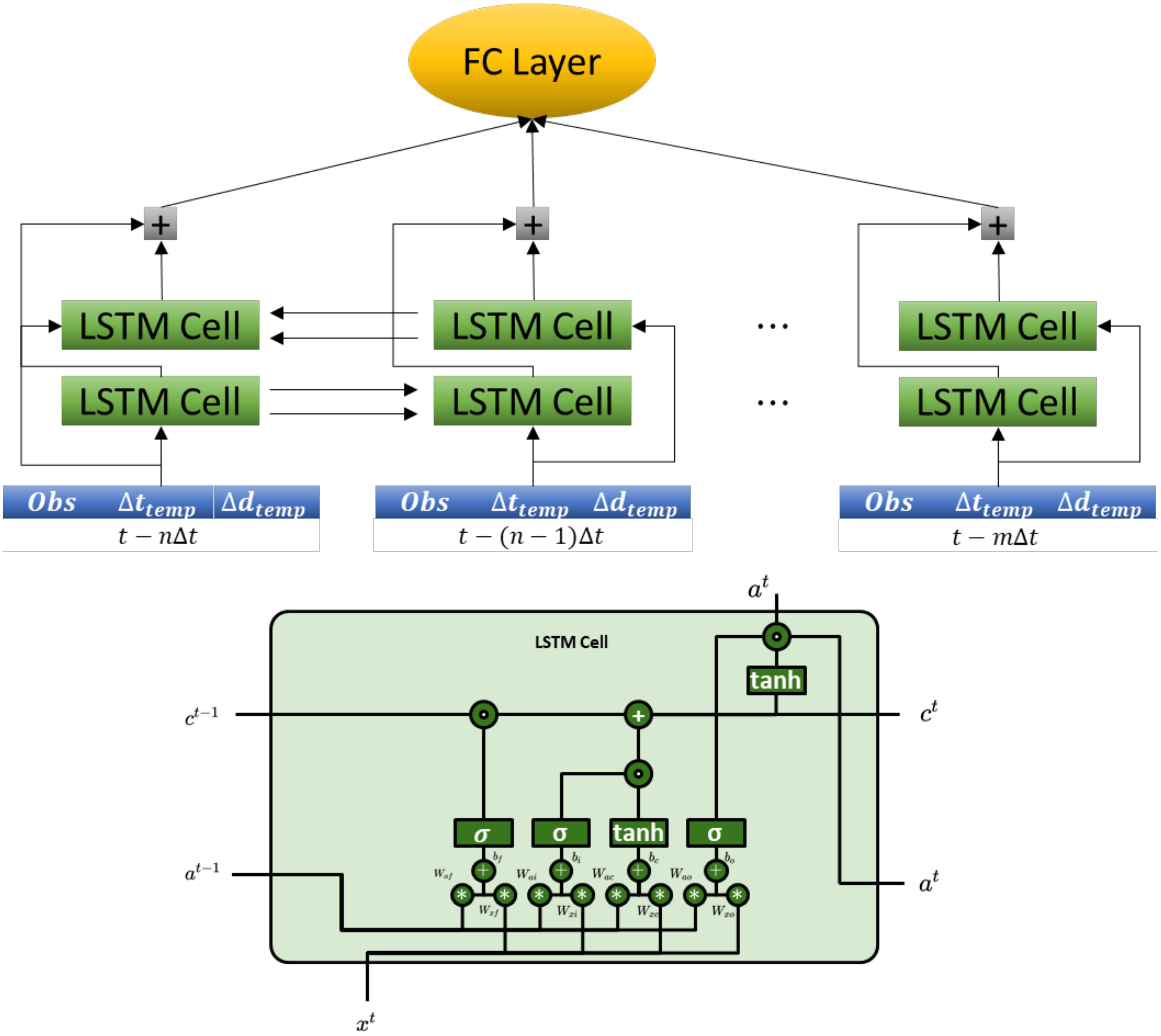

*Figure 7- Bidirectional LSTM architecture. LSTM cells are represented by rectangular units. At each timestep, the model is fed by features for all depths. The lower cells are associated with forward data while the upward cells represent backward-in-time propagation.*

### 2-4-E- Hyperparameter Tuning and Model Evaluation

The hyperparameters of a DL model must be optimized to produce the best possible inferences while avoiding overfitting. To compare the quality of DL methods, each must be tuned with their best hyper-parameter sets. For XGBOOST, the number of estimators, learning rate, maximum depth, and minimum child weight and regularization parameters must be tuned. The DL models used in this study have the following hyperparameters: number of layers, number of nodes at each layer, optimizer type and learning rate, dropout rate, and activation function. A plain-language explanation of these hyperparameters is provided in Table 2. Each model was tuned separately for each level of noise added to the input data. All of the DL methods use backpropagation to update the parameters of the network. In this study, we used the Adam algorithm, which is an extension of gradient descent. Adam computes individual adaptive learning rates for each of the network's parameters based on estimates of the first and second moments of the gradients (Kingma and Ba). All hyper-parameters (XGBOOST and DL) were tuned using a grid search approach. There is no agreement in the literature regarding which performance metrics should be used for different ML/DL algorithms in regression problems (Botchkarev) . We decided to use root mean squared error (RMSE) between the observed and model-calculated values of surface/ground water flux for training and assessment of model quality. The $R^2$ value was also calculated, but it was only used to further illustrate the quality of the predictions. Table 3 shows the best set of hyperparameters for each algorithm. To ensure that the performance of the algorithms was not dependent on the specific set of errors introduced or on the initial values assumed for the weights, we ran each combination of ML/DL method and noise filter 10 times using different seeds. The average performance and feature importance values are used for comparison among ML/DL methods.

*Table 2- Explanation of the hyperparameters tuned for each model type.*

| Algorithm | Hyperparameter | Definition |
|---|---|---|
| | Number of estimators | The number of weak learners |
| | Learning rate | Step size shrinkage used at each weak learner to prevent overfitting |
| **XGBOOST** | Maximum depth | Maximum depth (number of levels) of a tree. |
| | Subsample | The fraction of training data that is selected for training each tree |
| | Alpha | L1 regularization term on weights |
| | Minimum child weight | Minimum number of instances (samples) needed to justify forming a split |
| | Number of layers | The number of hidden layers in the deep learning algorithm |
| | Number of nodes | Number of computational units in a fully connected layer |
| **Deep Learning** | Learning rate | The fraction of the calculated gradient to be used to update the parameter values for a gradient descent iteration |
| | Dropout rate | The probability of setting a node to zero in the hidden layer to reduce overfitting |
| | Activation | A nonlinear function that transforms the summed weighted inputs of a node to an output value |

*Table 3- Optimized hyperparameter values for each ML model for noise-free, noisy, and filtered cases. Values are only shown for the best performing filter for each ML.*

| Model | Scenario | Filter | Hyperparameter |
|---|---|---|---|
| **DT** | Noise Free | | Maximum depth:15, min_samples_split: 0.001 |
| | Noisy | | Maximum depth:7, min_samples_split:0.01 |
| | Filtered | Flat | Maximum depth:7, min_samples_split:0.01 |
| **XGBOOST** | Noise Free | | Number of estimator:150, learning rate:0.1, max depth of learner:6, subsample:0.08, reg_alpha:0.1, min_child_weight:100, learning rate=0.1 |
| | Noisy | | Number of estimator:50, learning rate:0.3, max depth of learner:4, subsample:0.08, reg_alpha:0.1, min_child_weight:100, learning rate=1 |
| | Filtered | Hamming | Number of estimator:150, learning rate:0.1,max depth of learner:6, subsample:0.08, reg_alpha:0.1, min_child_weight:100, learning rate=1 |
| **MLP** | Noise Free | | Number of layer =2, hidden nodes layer 1:60 (relu), hidden nodes layer 2 :40 (relu), batch_size:1024, dropoutrate:0.5, learning rate=0.001 |
| | Noisy | | Number of layer :2, hidden nodes layer 1:60 (relu), hidden nodes layer 2:40 (relu), batch_size:512, dropoutrate:0.5, learning rate=0.001 |
| | Filtered | Flat | Number of layer =2, hidden nodes layer 1:60(relu), hidden nodes layer 2:40 (relu), batch_size:512, dropoutrate:0.5, learning rate=0.001 |
| **CNN** | Noise Free | | Number of conv layers:2, layer 1 number of filter:100 (relu), layer 2 number of filter:100 (relu), kernel size 1:4, kernel size2 :4, number of fully connected layer:2, Number of nodes fully connected 1:40 (sigmoid), Number of nodes fully connected 2:40 (sigmoid), dropoutrate:0.5, , learning rate=0.0001 |
| | Noisy | | Number of conv layers:2, layer 1 number of filter:150(relu), layer 2 number of filter:150 (relu), kernel size 1:4, kernel size2 :4, number of fully connected layer:2, Number of nodes fully connected 1:40 (sigmoid), Number of nodes fully connected 2:40 (sigmoid), dropoutrate:0.5, learning rate=0.0001 |
| | Filtered | Flat | Number of conv layers:2, layer 1 number of filter:80 (relu), layer 2 number of filter:80 (relu), kernel size 1:3kernel size2 :3, number of fully connected layer:2, Number of nodes fully connected 1:25 (sigmoid), Number of nodes fully connected 2:25(sigmoid), dropoutrate:0.5, learning rate=0.0001 |
| **LSTM** | Noise Free | | Number of hidden nodes:35, number of batches:128, lstm_type=Bi-LSTM, num_layers=1, learning rate =0.0008 |
| | Noisy | | Number of hidden nodes:45, number of batches:1024, lstm_type=Bi-LSTM, num_layers=1, learning rate =0.0008 |
| | Filtered | Bartlett | Number of hidden nodes:35, number of batches:256, lstm_type=Bi-LSTM, num_layers=1, learning rate =0.0008 |

## 2-5- Intrinsic and Extrinsic Measures of Feature Importance

The third objective of our study was to determine if existing tools can extract feature importance information from DL algorithms so that they could be used to improve monitoring network design. The first step in this analysis was to compare the feature importance provided intrinsically by tree-based methods with those identified by an independent analysis. Then, we compared the feature importances identified using the same independent analysis for different DL algorithms solving the same inference problem. Finally, we examined whether feature importance changed as a function of the level of added measurement error.

DL algorithms perform pattern recognition to model the relationship between input and output data. However, the complexity of some deep learning models and their large numbers of parameters, coupled with a lack of an underlying physics-based structure, can make it difficult to understand how information propagates through the model. For subsurface hydrology, which typically faces very restrictive monitoring budgets, measurement optimization that can lead to more efficient (less expensive) networks has real value. Unfortunately, the lack of transparency of deep learning algorithms can make it difficult to determine which inputs are informative for specific model predictions. This knowledge can, for instance, inform the choice between the adoption of more expensive sensors with lower measurement uncertainty versus a larger number of less accurate sensors, or the number of sample depths and placement of sensors for optimal model-aided interpretation.

Feature importance describes the relative contributions of the input features to the inference made using an ML/DL. In general, there are two approaches to infer feature importance: 1) intrinsic methods that rely on the ability of ML algorithm to extract feature importance during training; and 2) external approaches that operate on a trained DL, such as LIME (Ribeiro et al.) and ALE (Apley and Zhu). Tree-based methods have clear, intrinsic procedures to identify the contribution of each input feature to the final predictive model(Zheng et al.; Lai et al.). Until recently, this has been difficult to replicate for deep learning algorithms. But, some recent advances have proposed methods to estimate feature importance for neural networks (Apley and Zhu; Ribeiro et al.; Molnar et al.).

We used the Accumulated Local Effect (ALE) approach to extract the importance of each input feature for each combination of ML and noise filter (when applied). ALE illustrates how changes in features influence the output of a function (Devlin et al.). Specifically, ALE repeatedly predicts the target values with different subsets of the input features and then uses the difference of the predictions to determine the marginal importance of each features. The main effect ALE, $f_{i,ALE}$, of a feature, $x_i$, is defined as:

$$f_{i,ALE}(x_i) = \int_{x_{min,i}}^{x_i} E\left[f_i'(x_1, x_2, \dots x_p) \middle| X_i = z_i\right] dz_i - c_i \qquad \textit{(Equation-7)}$$

In which:

$$f_i'(x_1, x_2, \dots x_p) = \frac{\partial f(x_1, x_2, \dots x_p)}{\partial x_i} \qquad \textit{(Equation-8)}$$

The variable $x_{min,i}$ is min $\{x_{j,i}: i \in observation\ number\}$. In Eqs. [7&8], $f_i'$ calculates the local effects of $x_i = z_i$ on f () and these expected local effects are summed over all points in the vicinity of $X_i = z_i$. These local effects are accumulated over all values of $z_i$ up to our point of interest, $x_i$. Finally, a constant, $c_i$, is subtracted so that the mean of $f_{i,ALE}(x_i)$ with respect to the marginal distribution of $x_i$ is zero, which ensures that the ALE components sum to the full prediction function. The second order ALE interaction can be calculated in the same manner, but with a rectangular interval and conditioning over two variables, while adjusting for the main effect of features in addition to the overall mean effect. For this study, we used the variation in ALE main effects of the features at 10-quantiles of their distribution; a feature with higher variation in the ALE main effect indicates higher feature importance (Molnar).

To compare feature importances among DL algorithms, we first determined the aggregated importance of each feature type (temperature, temporal gradient, spatial gradient), then each measurement depth, and then each measurement time delay. For example, to calculate the importance of each feature type, we calculated the mean variation in ALE for each feature type, pooling the importance of each measurement type over all depths and time delays. Similar pooling was done to isolate measurement depths and time delays.

Another approach to comparing feature importance among DL algorithms is to compare the similarity of their feature importance rankings using the Jaccard Index (Lai et al.). This distance measure compares the number of common elements in two sets to the union of all features in both sets (Eq. 9). We first ranked the features for each model based on the ALE results and selected the top K ranked features. We calculated the Jaccard Index between the top K ranked features, TopK(), for each pair of DL models. We then estimated the similarity between the DL models as:

$$\text{Similarity}\ (\text{m}, \text{m'}) = \frac{\left|TopK\left(I_f^m\right) \cap TopK\left(I_f^{m'}\right)\right|}{\left|TopK\left(I_f^m\right) \cup TopK\left(I_f^{m'}\right)\right|} \qquad \textit{(Equation-1)}$$

In which, m is the model, f is the feature, and $I_f^m$ is the ALE variation for a given feature and model.

# 3- Results and Discussion

The first two goals of this study were to test the ability of ML/DL methods to infer surface/ground water flux with high temporal resolution from subsurface temperature observations with and without noise. Our third goal was to determine whether meaningful feature importance information could be retrieved from the trained DL models. In this section, the ability of each ML/DL model is first assessed based on its ability to infer surface/ground water flux given noise-free synthetic data. Then, to assess the resilience of each ML/DL model to measurement errors, they were retrained with different levels of noise applied to the temperature time series. Finally, we examined the effectiveness of five classic noise filtering techniques (Hamming, Blackman, Hanning, Bartlett and flat) to improve the performance of each ML/DL model given noisy data. For each stage of the analyses, feature importance analysis was conducted to identify the most informative observations.

## 3-1- Base Case Evaluation Using XGBOOST

To bridge from the examination of simpler, tree-based ML models (M. Moghaddam et al.) to DL approaches, we first tested the performance of XGBOOST. For noise-free synthetic data, considering both upward and downward flux, XGBOOST performed well, showing an RMSE of $8.96 \times 10^{-7}$ ($R^2$= 0.981) and $3.99 \times 10^{-6}$ ($R^2$=0.795) for the training and testing sets, respectively (Figure 8A, B). When noise with SNR=100 was added, the RMSE of the training and testing increased to $4.18 \times 10^{-6}$ ($R^2$= 0.57) and $7.89 \times 10^{-6}$($R^2$= 0.13) (Figure 8C, D). This result is consistent with (M. Moghaddam et al.), who showed that tree-based algorithms have a clear decrease in performance when subjected to noise, especially when estimating downward flux (shown on (Figure 8) as negative upward flux). The most successful noise filtering approach for XGBOOST was Hamming. With this filter applied, the RMSE of the training and testing sets were $1.66 \times 10^{-6}$ 1 ($R^2$=0.935) and $5.36 \times 10^{-6}$ 5.( $R^2$=0.59), respectively (Figure 8E, F). This level of performance on the testing set is still likely to be unacceptable for most hydrologic applications.

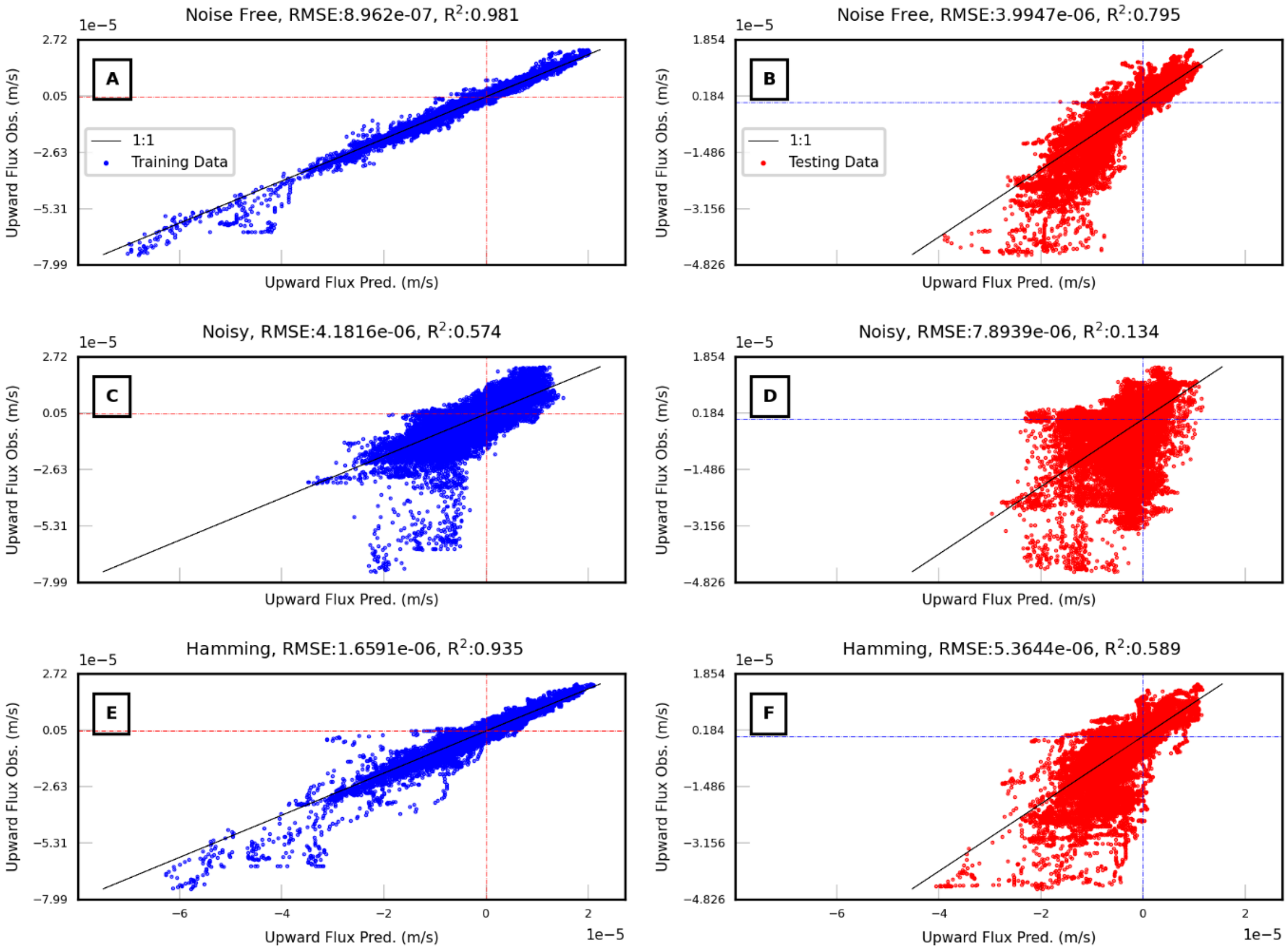

*Figure 8- Training (left) and testing (right) results of XGBOOST using temperature sensors, which are located at, 0.005, 0.15 m, 0.255 m and 1.995 m. A, B: fits to noise-free data .C, D: fits to noisy data with SNR=100. E, F: fits to filtered data using the best filter performing (flat).*

## 3-2- Evaluation of Deep Learning Algorithms

Identical analyses were performed on XGBOOST and the three DL models (Figure 9). Detailed results are only shown for CNN, which was found to perform best overall among the DL models. (A more detailed comparison is provided in the Discussion.)

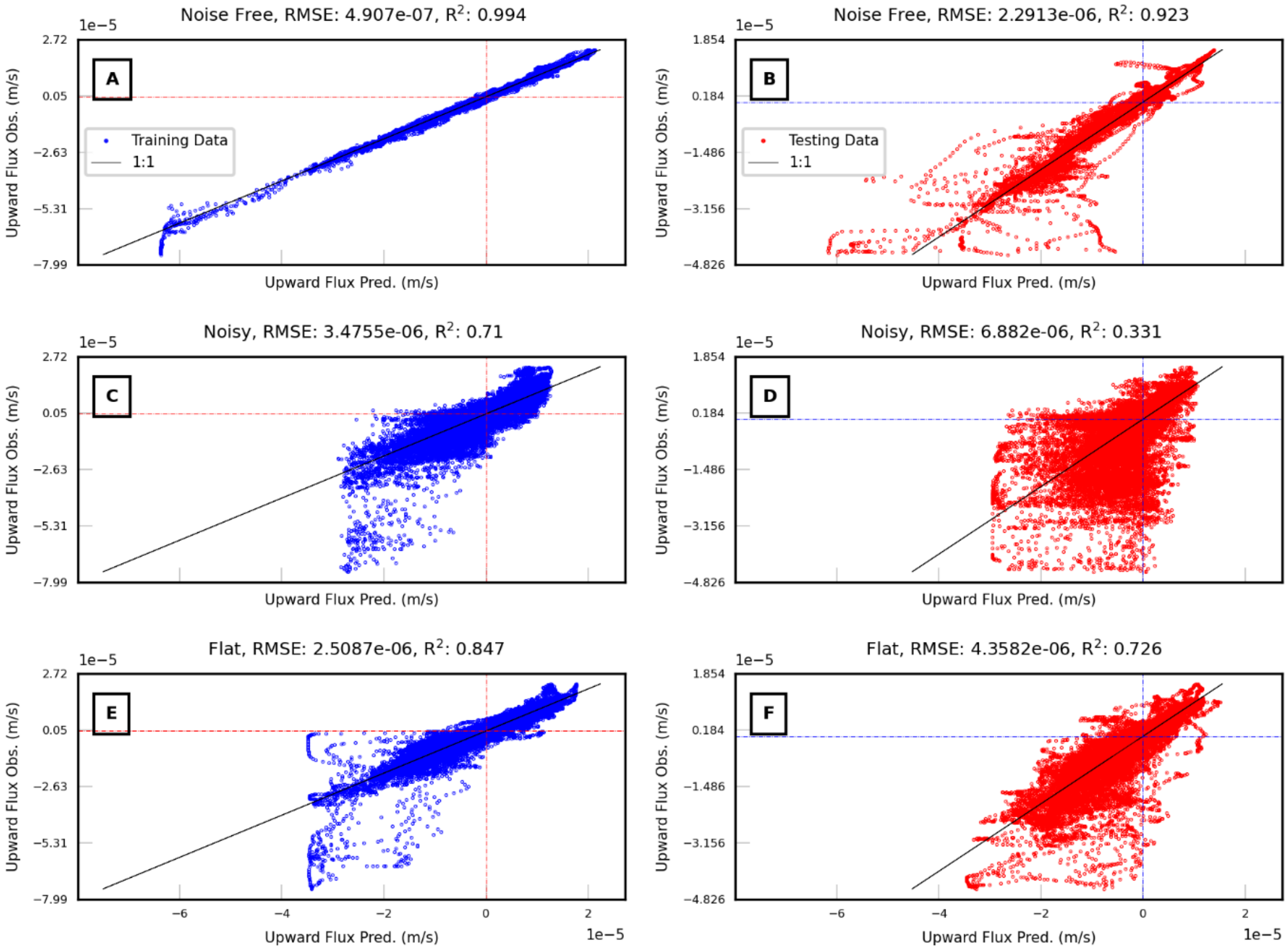

*Figure 9- Training (left) and testing (right) results for CNN trained on temperature observations, only, located at, 0.005, 0.15 m, 0.255 m and 1.995 m depth. A, B: fits for noise-free data .C, D: fits for noisy data with SNR=100. E, F: fits for filtered data using the best-performing filter (flat).*

As expected, there is higher error for testing than training sets and there is increasing error transitioning from no noise, to filtered, to noisy conditions. For noise-free synthetic data, CNN performed very well, showing an RMSE of $4.9 \times 10^{-7}$ ($R^2$=0.99) and $2.29 \times 10^{-6}$ (R2 = 0.92) for the training and testing sets, respectively (Figure 9A, B). When noise with SNR=100 was added, the RMSE of training and testing increased to $3.47 \times 10^{-6}$ (R2 = 0.71) and $6.88 \times 10^{-6}$ ($R^2$=0.33), respectively (Figure 9C, D). The flat filter was most successful for noise removal for CNN, leading to an RMSE of the training and testing sets of $2.51 \times 10^{-6}$ ($R^2$=0.85) and $4.36 \times 10^{-6}$ ($R^2$=0.73), respectively (Figure 9E, F).

It is difficult to say whether this level of performance on the filtered noisy data would be acceptable for hydrologic applications. To put this performance in context, we compared these results with those shown on Figure 8 of (Lautz). Given the different flux ranges in the two studies, we scaled the RMSE by the range of fluxes reported in each study:

$$RMSR_N = \frac{RMSE}{flux_{max} - flux_{min}} \qquad \textit{(Equation-10)}$$

(Lautz) showed RMSEN values of 0.27 and 0.18 for unfiltered and filtered cases, respectively. Our application of CNN with and without a flat filter showed RMSEN values of 0.10 and 0.06, respectively, suggesting that the CNN-based analysis of filtered, noisy time series would be acceptably accurate for hydrologic applications.

## 3-3- Detailed Comparisons of the Performance of the Deep Learning Algorithms

The performance of the ML models for noise-free, noisy (SNR=100), and filtered temperature data can be compared across all combinations of ML model and noise filter (Figure 10 A,B,C). The RMSE value is shown and the values are highlighted by the background color (darker green indicates worse performance). Similar to (M. Moghaddam et al.), we found significant differences in performance of both the tree-based method and the DL models for upward and downward flow (Figure 10 A, B). The DL algorithms perform better than the simpler tree-based methods. CNN showed the best performance for noise-free and filtered noisy data, which is most likely to represent the most common mode of application to real data. CNN consistently outperforms the other ML models, but the choice of best filter for CNN differs for upward and downward flow. To ensure that the ML performance did not depend on the specific error realization, the performance was repeated for ten temperature observation error realizations. The variance of the RMSE for each ML over the multiple restarts, shown as error bars on (Figure 10D), is small compared to the differences in performance among models. Finally, the superior performance of CNN for filtered (denoised) data persisted over noise levels ranging from SNR=100 to 4000 (Figure 11).

## 3-4- Comparison of Accuracy of Inferring Upward versus Downward Flow

For all scenarios, models, and filter types, there was better performance when resolving upward flux compared to downward flux (Figure 10). Our findings, support those of (M. Moghaddam et al.), but they differ from our expectations, based on experience with using numerical and analytical models to interpret temperature profiles [e.g. Soto-López et al., 2011]. For those applications, downward flux is easier to infer than upward because these models rely on advective heat transport to carry a surface temperature signal to depth as the basis for inferring water flux. For this application, flow is 3D on a larger scale; when flow is upward, the temperature signal at depth is not exclusively driven by temperature changes occurring at the ground surface where flux is being estimated. Rather, in some cases, temperature variations at the bottom boundary have a controlling influence on the temperature time series at the observation depths. This condition violates the assumptions of most analytical and numerical model analyses, making it difficult or impossible to interpret upward flux. However, one potential advantage of an ML/DL-based analysis is that the interpretations are model-free. Therefore, they may be able to find patterns associated with upward flow.

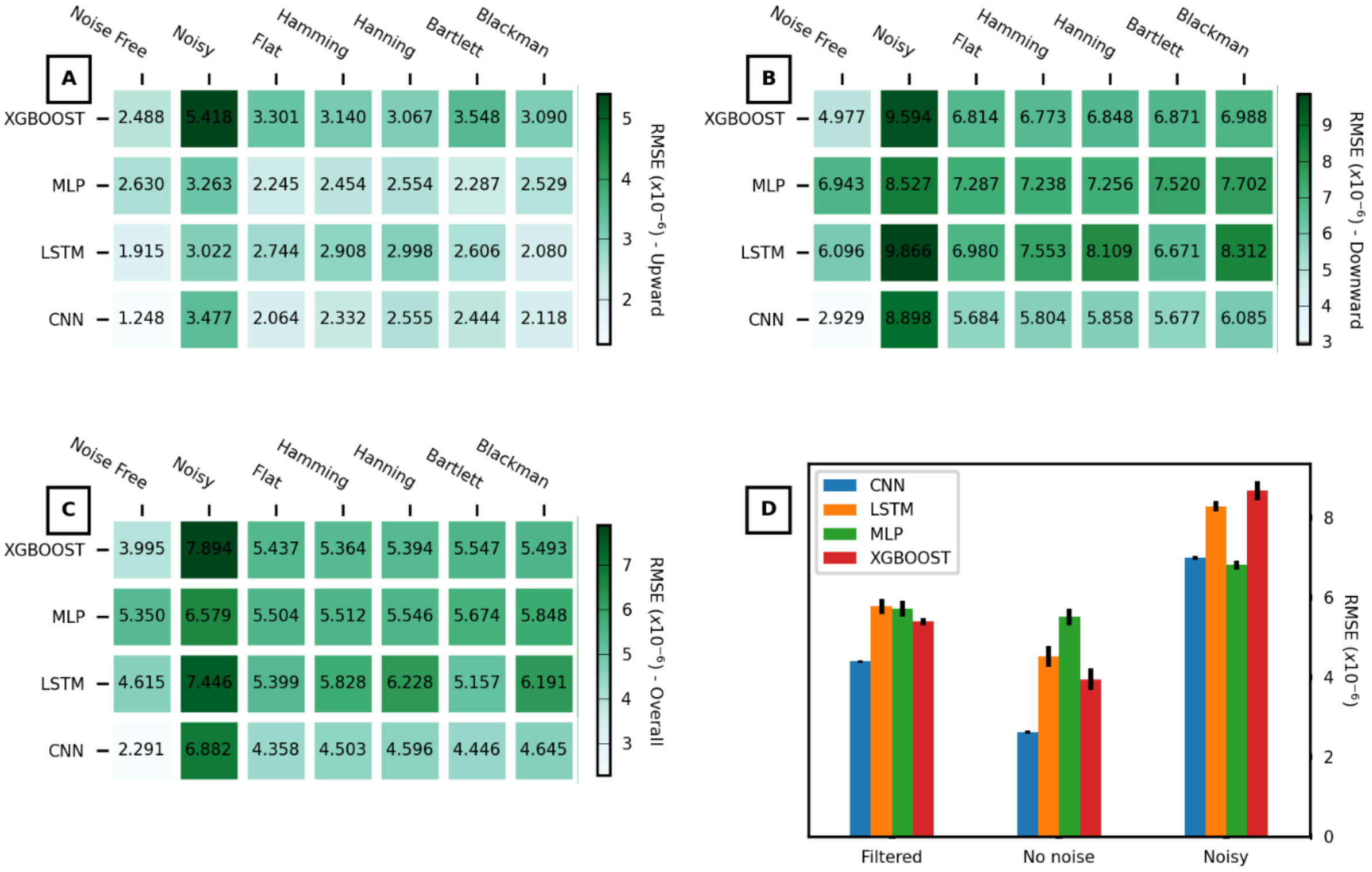

*Figure10- RMSE between prediction and observation for each ML method for noise-free, noisy, and filtered observations. A: only upward flow; B: only downward flow; and C: all flow conditions; D: Performance of the machine learning algorithms for noise-free, noisy, and filtered temperature observations subjected to best performed filters for filtered scenario. The variance over ten restarts is shown as error bars.*

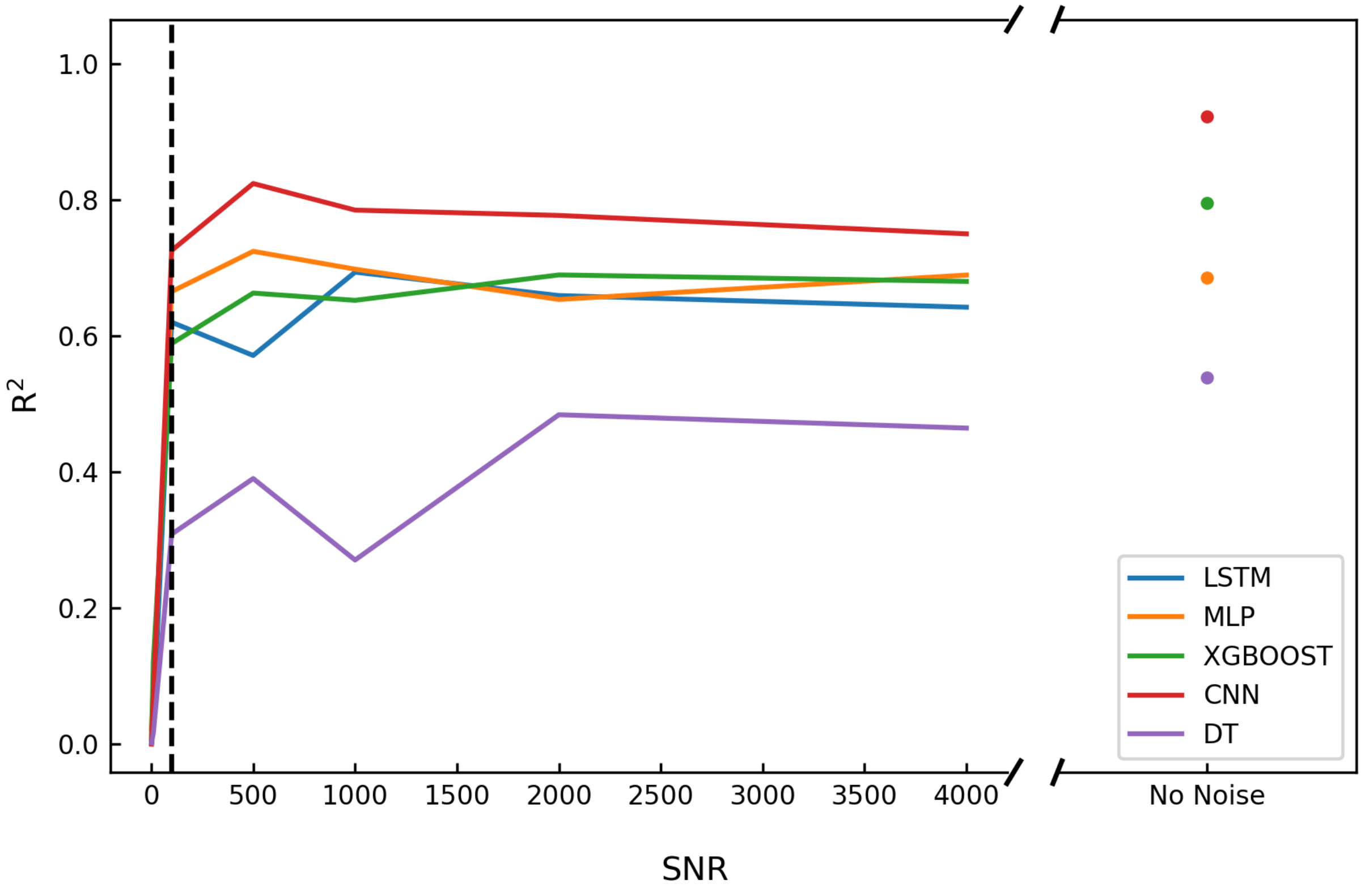

*Figure 11- Performance of MLs on denoised data as a function of the SNR using the best noise filter for each ML at each noise level. Dot points at the left shows the performance of model for noise free scenario*

Close examination of the subsurface conditions related to upward and downward flow are illustrated through consideration of the temperature profiles collected around three times, shown on the surface/ground water flux times series (Figure 12 A). Two of these times experience downward flux (one high at time step 87925 and one intermediate at time step 30125) and the third shows an intermediate upward flux at time step 39580. Figure 12 B, C, D shows the surface/ground water flux time series within ±1000 time steps of the identified times. The time series are similar for intermediate flow conditions for both downward (Figure 12A) and upward (Figure 12B) flow; there is considerable high frequency variability combined with relatively large changes in flux over 125 time steps (approximately ten hours). This time scale of variation in surface flux magnitude, together with the stated aim of inferring the flux with high temporal resolution, supports the consideration of a relatively short time window on the order of one hour.

The responses during high downward flux (Figure 12D) are associated with smoothly varying, isolated events, which are orders of magnitude larger and separated by approximately 20 hours. Figures 12 E, F, G show the flux time series within ±100 time steps of the identified times and Figures 12 H, I, J show the corresponding temperature time series at the ground surface and at the bottom boundary of the domain. Both upward and downward events vary relatively smoothly over approximately 100 time steps (~8 hours). There are no immediately obvious temporal patterns in the temperature top and bottom boundary conditions that indicate why upward flux (center column) is simpler to infer than downward flux (left and right columns). There is a relatively high degree of similarity of the boundary temperature time series for upward and downward flux. In contrast, the temperature profiles are significantly different for the two flow directions (Figure 12 K, L, M). Specifically, the temperature profile is smoothly varying over the entire profile with relatively low local gradients for downward flux conditions (Figure 12 K, M); whereas the temperature variations are highly localized at the near surface for upward flow conditions (Figure 12L). As an example, Figure 12 K shows temperature profiles every ten time steps from -100 to 100 about the time of flux estimation. Figure 12H shows that the surface temperature is 9.2 $^{o}$C 100 time steps previous to flux estimation. Figure 12H shows the 9.2 $^{o}$C isotherm progressing to progressively deeper depths through time. In contrast, Figure 12J, which is also associated with downward flux, shows very different responses. There is a progression of a low temperature at the surface to depth, but this signal is mixed with a very strong signal associated with displacing the previous temperature profile. Finally, the upward flow profile (Figure 12I) is entirely different. There is a very sharp temperature gradient with depth, but the influence of the upward flow is to displace the pre-existing temperature profile upward. As a result, the upward flow signal benefits from very high spatial gradients while suffering from much less mixture of temperature signals from multiple flow periods. In this case, the temperature profiles are consistent with our expectation that downward flux would communicate surface temperature variations to greater depths than upward flux. This was perceived as a limitation from the point-of-view of model-based interpretation. But, interpreting the profiles in a model-free, ML context turns these characteristics into an advantage, leading to better performance for inferring upward rather than downward flow.

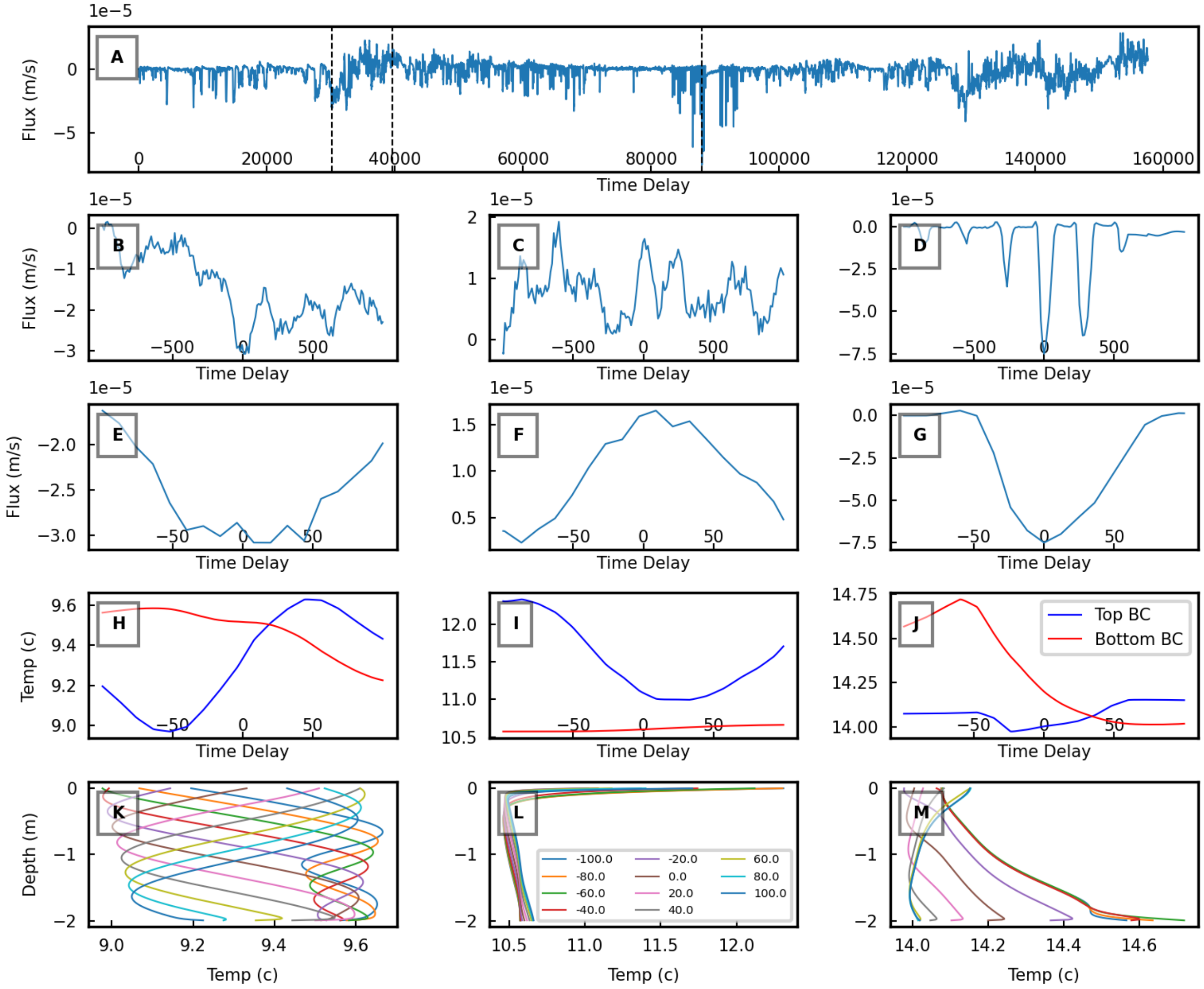

*Figure 12- Temperature and flux profiles for different conditions. A: Selected upward and downward flux times on flux time series. B, C, D: the surface/ground water flux time series within ±1000 time steps at 30125, 39580, 8792, respectively. E, F, G: the surface/ground water flux time series within ± 100 time steps at 30125, 39580, 8792, respectively. H, I, J: The boundary temperature time series for upward and downward at 30125, 39580, 8792 time steps respectively. K, L, M: Temperature profile at 30125, 39580, 8792.*

## 3-5- Feature Importance Analyses

The DL methods do not include an intrinsic measure of feature importance; therefore, we used ALE to assess which inputs were most useful for inferring ground/surface water flux. To test the validity of this approach, we first compared the feature importance rankings of XGBOOST using its intrinsic method to those determined using ALE. The XGBOOST built-in function ranks features based on expected reduction in node impurity while ALE calculates the variance of change in output prediction across all possible values of a feature. Therefore, the absolute values of the metrics are different; but, generally, only relative rankings are used to assess the relative importance of observations. We first assessed the feature importance based on the ±30-minute time window used for the previous analyses (Figure 13). The very high importance of the extreme observation times for both the built-in and ALE analyses could have indicated that even later (or earlier) times would provide more information for constraining the exchange flux. (The high importance of the earliest times is concerning as it has no physical basis.) To examine this, we extended the positive time window to 60 time steps (5 hours), which is well beyond any physically reasonable time delay for the highly dynamic conditions examined. The results for the built-in analysis (Figure 13C) show no consistent pattern for the optimal time delay. This may indicate that the optimal delay depends on the flux or that more complicated elements of the flux and temperature sequences make it impractical to down-sample the delay time. Practically, this suggests that many time delays should be considered in the ML analyses, which is relatively simple to implement and does not impose additional observation or analysis costs. In contrast, the ALE-based feature importance analysis continues to indicate the highest importance for the extreme positive and negative delays. These results are not physically reasonable, suggesting that ALE is not suitable for assessing optimal measurement times. This failure may be due to the high correlation of temperature in time (Molnar et al.), but this cannot be confirmed based on our results. In contrast, the importance of different feature types and locations based on the built-in functions were not affected by extending the time window (not shown). Therefore, we based the following discussions on the more physically realistic ±30 minute window used for the previous analyses. The built-in function and ALE agree that temperature measurements are more important than spatial or temporal gradients (Figure 13E, F). Both methods show a slight preference to include a deep observation, although they disagree on the specific depth of the shallower observation (Figure 13G, H). However, no observation depth showed low enough importance to warrant reducing the monitoring network for XGBOOST-based analyses.

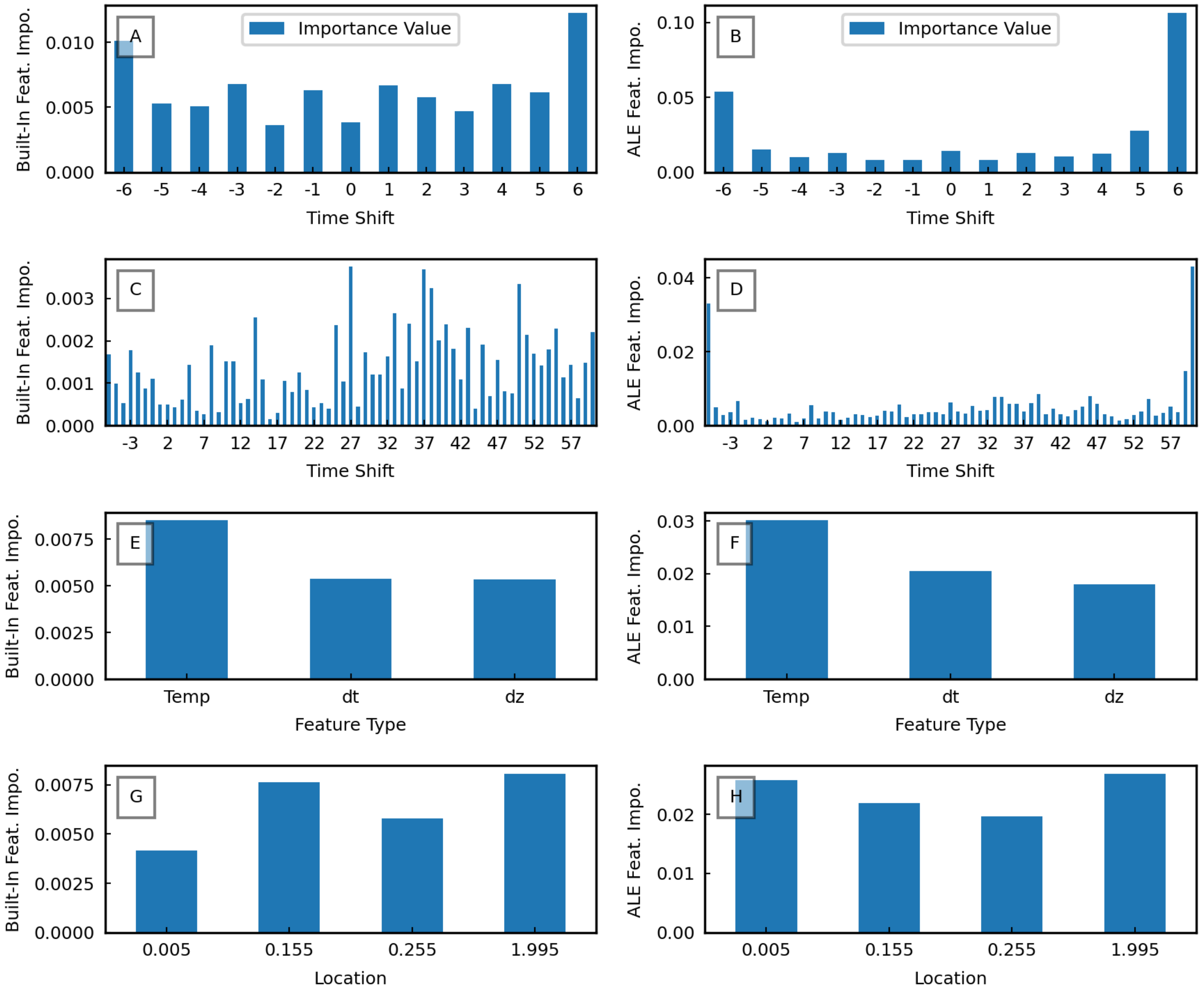

*Figure 13- Left column: feature importance analyses based on the built-in XGBOOST method; right column: feature importance based on ALE analyses. A, B,: feature importance between -6 and 6 time steps; C,D: feature importance between -6 and 60 time steps, E,F: feature importance by type – observed temperature, temporal gradient (dt), and spatial gradient (dz) for time shifts of -6 and 6, G,H: importance by depth between -6 and 6 time steps.*

Given the modest agreement between ALE and the built-in function for XGBOOST, feature importance analyses based on ALE for the DL methods must be considered with some caution. Given that warning, we examined the feature importance for each DL method when applied to noisy data with its highest performing noise filter. Given that the built-in XGBOOST method found no clear optimal measurement delays, we only compare the importance of features by type and depth from -6 to 6 times steps (Figure 14). Unlike XGBOOST, the DL methods favored temporal gradients over direct temperature observations (Figure 14A). But, similar to XGBOOST, there was not a strong enough preference for specific observation depths to support measurement network reduction (Figure 14B).

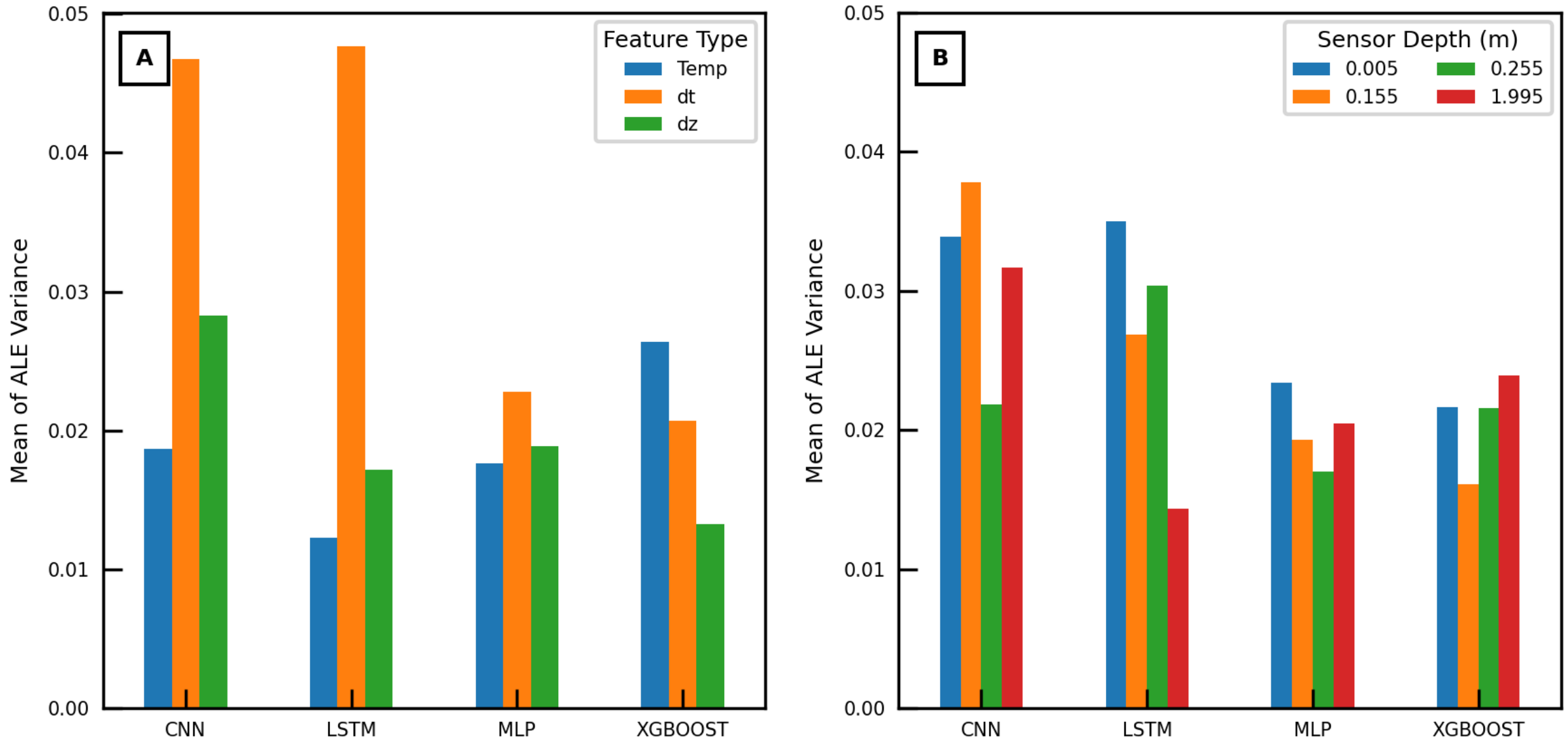

*Figure 14- feature importance comparisons for deep learning models for time shifts of -6 and 6. A: Feature importance by feature type; B: feature importance by depth*

A more detailed analysis of the ALE results considered each individual observation (combination of time delay, type, and depth) and ranked them by importance for each DL method using its best-performing filter. Once the features were listed by decreasing importance for each ML, we compared the similarities of the top third (46) most important features among them (Figure 15). The agreement is quite low; the DL methods only share between 15 and 30% of their most informative features.

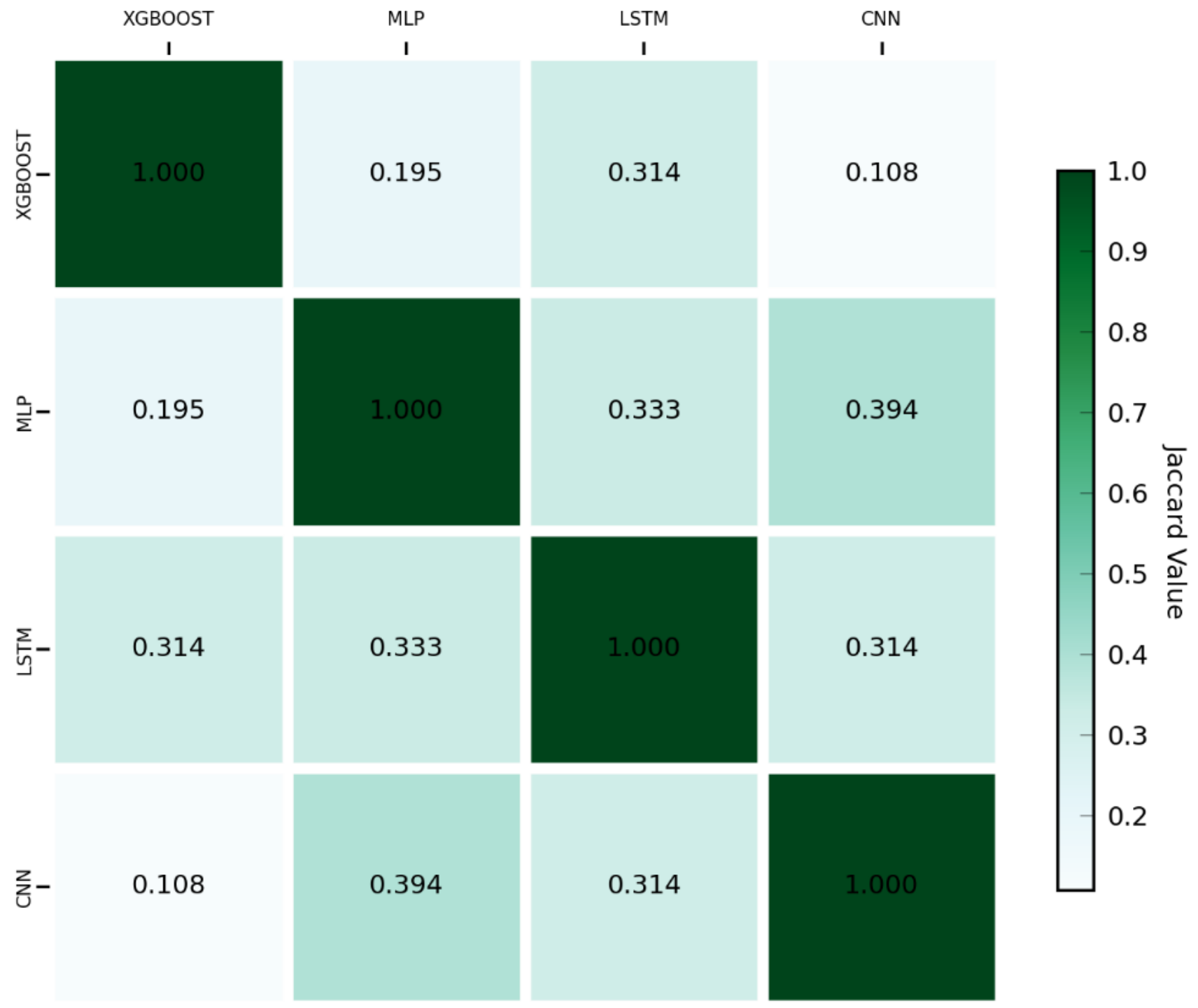

*Figure 15- Jaccard similarity between the top 46 features of different models based on ALE method. The similarity is the greatest between XGBOOST and RT.*

## 4- Conclusions

The present study investigated the capabilities and limitations of DL algorithms. Initially, we attempted to infer surface/ground water flux from subsurface temperature observations with high temporal resolution (5 min) in a highly dynamic system. We found that the DL methods were more accurate than the tree-based ML methods, especially for noisy data with noise filtering applied. We repeated our analyses with different levels of added observational noise with and without noise filtering applied to the corrupted data. Among the DL methods, CNN was most successful for all levels of added noise. However, the differences in absolute performance among DL methods and between DL and ML methods were relatively small. Given the considerably shorter run times and built-in feature importance assessment, XGBOOST may be preferable for practical applications of the examined methods for estimation of surface/ground water exchange flux and for associated monitoring network optimization efforts (M. Moghaddam et al.). We were surprised to find that upward flux was predicted more accurately than downward flux. After examining the temperature profiles for both conditions, we conclude that this downward flux conditions suffered from lower temperature gradients with depth and were more likely to have mixed signals from current and past conditions than upward flux conditions, leading to this difference in performance. This result is in direct contrast to previous investigations using numerical models to infer surface/ground exchange flux and it may indicate that combined use of ML or DL inference with numerical inference could improve flux estimation beneath river systems.

A secondary goal of this study was to use the Accumulated Local Effect method to examine whether some observations could be eliminated from the data set because they do not provide information to the DLs. Our results found that the DLs extracted information from all of the observations. This suggested that observation networks could not be reduced without loss of fidelity. This result may also indicate that the DL methods do not tend to localize information, potentially making them less useful for measurement network optimization than simpler tree-based methods. In summary, we found that the DL methods showed real promise for inferring surface/ground exchange flux from subsurface temperature methods. However, it is unclear whether the increased complexity and lack of clarity regarding feature importance of the DLs compared to the simpler MLs justifies their use for all applications.

## Acknowledgement

This research was supported by the U.S. Department of Energy (DOE), Office of Biological and Environmental Research (BER), as part of Environmental System Science (ESS) program. This contribution originates from the River Corridor Scientific Focus Area (SFA) at the Pacific Northwest National Laboratory (PNNL). PNNL is operated for the DOE by Battelle Memorial Institute under contract DE-AC05-76RL01830. This research used resources of the National Energy Research Scientific Computing Center, a DOE Office of Science User Facility supported by the Office of Science of the U.S. Department of Energy under contract DE-AC02-05CH11231. This paper describes objective technical results and analysis. Any subjective views or opinions that might be expressed in the paper do not necessarily represent the views of the U.S. Department of Energy or the United States Government.

## Data Availability Statement

The data that supports the findings of this study are openly available in Center For Research Data at https://doi.org/10.4121/uuid:c874327b-9d70-4fd4-b96c-478eebd8ba21. This material was prepared as an account of work sponsored by an agency of the United States Government. Neither the United States Government nor the United States Department of Energy, nor Battelle, nor any of their employees, nor any jurisdiction or organization that has cooperated in the development of these materials, makes any warranty, express or implied, or assumes any legal liability or responsibility for the accuracy, completeness, or usefulness or any information, apparatus, product, software, or process disclosed, or represents that its use would not infringe privately owned rights. Reference herein to any specific commercial product, process, or service by trade name, trademark, manufacturer, or otherwise does not necessarily constitute or imply its endorsement, recommendation, or favoring by the United States Government or any agency thereof, or Battelle Memorial Institute. The views and opinions of authors expressed herein do not necessarily state or reflect those of the United States Government or any agency thereof.

## Declaration of Competing Interest

The authors declare that they have no known competing financial interests or personal relationships that could have appeared to influence the work reported in this paper.

## References

Adhikari, Abishek, et al. "Comparative Assessment of Snowfall Retrieval From Microwave Humidity Sounders Using Machine Learning Methods." *Earth and Space Science*, vol. 7, no. 11, Nov. 2020, doi:10.1029/2020EA001357.

Afzaal, Hassan, et al. "Groundwater Estimation from Major Physical Hydrology Components Using Artificial Neural Networks and Deep Learning." *Water (Switzerland)*, vol. 12, no. 1, 2020, doi:10.3390/w12010005.

Aldhaif, Abdulmonam M., et al. "Characterization of the Real Part of Dry Aerosol Refractive Index Over North America From the Surface to 12 Km." *Journal of Geophysical Research: Atmospheres*, vol. 123, no. 15, Blackwell Publishing Ltd, Aug. 2018, pp. 8283–300, doi:10.1029/2018JD028504.

Aliper, Alexander, et al. "Deep Learning Applications for Predicting Pharmacological Properties of Drugs and Drug Repurposing Using Transcriptomic Data." *Molecular Pharmaceutics*, vol. 13, no. 7, American Chemical Society, July 2016, pp. 2524–30, doi:10.1021/acs.molpharmaceut.6b00248.

Angelaki, Anastasia, et al. "Estimation of Models for Cumulative Infiltration of Soil Using Machine Learning Methods." *ISH Journal of Hydraulic Engineering*, Taylor and Francis Ltd., 2018, doi:10.1080/09715010.2018.1531274.

Angrisani, Leopoldo, et al. "PSD Estimation in Cognitive Radio Systems: A Performance Analysis." *19th IMEKO TC4 Symposium - Measurements of Electrical Quantities 2013 and 17th International Workshop on ADC and DAC Modelling and Testing*, 2013, pp. 543–48.

Anibas, Christian, et al. "Transient or Steady-State? Using Vertical Temperature Profiles to Quantify Groundwater-Surface Water Exchange." *Hydrological Processes*, vol. 23, no. 15, 2009, pp. 2165–77, doi:10.1002/hyp.7289.

Apley, Daniel W., and Jingyu Zhu. *Visualizing the Effects of Predictor Variables in Black Box Supervised Learning Models*. Dec. 2016, http://arxiv.org/abs/1612.08468.

Arabzadeh, Alireza, et al. "Global Intercomparison of Atmospheric Rivers Precipitation in Remote Sensing and Reanalysis Products." *Journal of Geophysical Research: Atmospheres*, vol. 125, no. 21, Nov. 2020, doi:10.1029/2020JD033021.

Assem, Haytham, et al. "Urban Water Flow and Water Level Prediction Based on Deep Learning." *Lecture Notes in Computer Science (Including Subseries Lecture Notes in Artificial Intelligence and Lecture Notes in Bioinformatics)*, vol. 10536 LNAI, 2017, pp. 317–29, doi:10.1007/978-3-319-71273-4_26.

Becker, MW, et al. "Estimating Flow and Flux of Ground Water Discharge Using Water Temperature and Velocity." *Journal of Hydrology*, 2004, pp. 221–33, https://www.sciencedirect.com/science/article/pii/S0022169404001921.

Bishop, Chris M. "Training with Noise Is Equivalent to Tikhonov Regularization." *Neural Computation*, vol. 7, no. 1, MIT Press - Journals, Jan. 1995, pp. 108–16, doi:10.1162/neco.1995.7.1.108.

Bisht, Gautam, et al. "Coupling a Three-Dimensional Subsurface Flow and Transport Model

with a Land Surface Model to Simulate Stream-Aquifer-Land Interactions (CP v1.0)." *Geoscientific Model Development*, vol. 10, no. 12, 2017, pp. 4539–62, doi:10.5194/gmd-10-4539-2017.

Botchkarev, Alexei. "A New Typology Design of Performance Metrics to Measure Errors in Machine Learning Regression Algorithms." *Interdisciplinary Journal of Information, Knowledge, and Management*, vol. 14, Informing Science Institute, Sept. 2019, pp. 45–76, doi:10.28945/4184.

Bouktif, Salah, et al. "Optimal Deep Learning LSTM Model for Electric Load Forecasting Using Feature Selection and Genetic Algorithm: Comparison with Machine Learning Approaches." *Energies*, vol. 11, no. 7, MDPI AG, June 2018, p. 1636, doi:10.3390/en11071636.

Briggs, Martin A., et al. "Using High-Resolution Distributed Temperature Sensing to Quantify Spatial and Temporal Variability in Vertical Hyporheic Flux." *Water Resources Research*, vol. 48, no. 2, 2012, doi:10.1029/2011WR011227.

Chakraborty, Supriyo, et al. "Interpretability of Deep Learning Models: A Survey of Results." *2017 IEEE SmartWorld Ubiquitous Intelligence and Computing, Advanced and Trusted Computed, Scalable Computing and Communications, Cloud and Big Data Computing, Internet of People and Smart City Innovation, SmartWorld/SCALCOM/UIC/ATC/CBDCom/IOP/SCI 2017* - , 2018, pp. 1–6, doi:10.1109/UIC-ATC.2017.8397411.

Chang, Haibin, and Dongxiao Zhang. "Machine Learning Subsurface Flow Equations from Data." *Computational Geosciences*, vol. 23, no. 5, Springer International Publishing, Oct. 2019, pp. 895–910, doi:10.1007/s10596-019-09847-2.

Chen, Kewei, et al. "Using Ensemble Data Assimilation to Estimate Transient Hydrologic Exchange Fluxes under Highly Dynamic Flow Conditions." *AGUFM*, vol. 2019, 2019, pp. H12I-20.

Chen, Tianqi, and Carlos Guestrin. "XGBoost: A Scalable Tree Boosting System." *Proceedings of the ACM SIGKDD International Conference on Knowledge Discovery and Data Mining*, vol. 13-17-Augu, Association for Computing Machinery, 2016, pp. 785–94, doi:10.1145/2939672.2939785.

Chen, Tianqi, and Tong He. "Xgboost : EXtreme Gradient Boosting." *R Package Version 0.4-2*, 2015, pp. 1–4, http://cran.fhcrc.org/web/packages/xgboost/vignettes/xgboost.pdf.

Coley, Connor W., et al. "A Graph-Convolutional Neural Network Model for the Prediction of Chemical Reactivity." *Chemical Science*, vol. 10, no. 2, 2019, pp. 370–77, doi:10.1039/c8sc04228d.

Constantz, Jim. "Heat as a Tracer to Determine Streambed Water Exchanges." *Water Resources Research*, vol. 46, no. 4, Blackwell Publishing Ltd, Apr. 2008, doi:10.1029/2008WR006996.

Cuthbert, M. O., and R. MacKay. "Impacts of Nonuniform Flow on Estimates of Vertical Streambed Flux." *Water Resources Research*, vol. 49, no. 1, Blackwell Publishing Ltd, 2013, pp. 19–28, doi:10.1029/2011WR011587.

Dadashazar, Hossein, et al. “Stratocumulus Cloud Clearings: Statistics from Satellites, Reanalysis Models, and Airborne Measurements.” *Atmospheric Chemistry and Physics*, vol. 20, no. 8, 2020, pp. 4637–65, doi:10.5194/acp-20-4637-2020.

de Vito, Laurent. “LinXGBoost: Extension of XGBoost to Generalized Local Linear Models.” *Arxiv.Org*, 2017, https://arxiv.org/abs/1710.03634.

Devlin, Summer, et al. *Disentangled Attribution Curves for Interpreting Random Forests and Boosted Trees*. May 2019, http://arxiv.org/abs/1905.07631.

Ehsani, Mohammad Reza, Jorge Arevalo, et al. “2019–2020 Australia Fire and Its Relationship to Hydroclimatological and Vegetation Variabilities.” *Water*, vol. 12, no. 11, Nov. 2020, p. 3067, doi:10.3390/w12113067.

Ehsani, Mohammad Reza, Ali Behrangi, et al. “Assessment of the Advanced Very High-Resolution Radiometer (AVHRR) for Snowfall Retrieval in High Latitudes Using CloudSat and Machine Learning.” *Journal of Hydrometeorology*, vol. 22, no. 6, American Meteorological Society, Apr. 2021, pp. 1591–608, doi:10.1175/jhm-d-20-0240.1.

Ehsani, Mohammad Reza, Ariyan Zarei, et al. *Nowcasting-Nets: Deep Neural Network Structures for Precipitation Nowcasting Using IMERG*. Aug. 2021.

Ehsani, Mohammad Reza, and Ali Behrangi. *On the Importance of Gauge-Undercatch Correction Factors and Their Impacts on the Global Precipitation Estimates*. no. June, 2021, doi:10.20944/preprints202106.0179.v1.

Essaid, Hedeff I., et al. “Using Heat to Characterize Streambed Water Flux Variability in Four Stream Reaches.” *Journal of Environmental Quality*, vol. 37, no. 3, Wiley, May 2008, pp. 1010–23, doi:10.2134/jeq2006.0448.

Georganos, Stefanos, et al. “Very High Resolution Object-Based Land Use-Land Cover Urban Classification Using Extreme Gradient Boosting.” *IEEE Geoscience and Remote Sensing Letters*, vol. 15, no. 4, 2018, pp. 607–11, doi:10.1109/LGRS.2018.2803259.

Gupta, Hoshin V., et al. “Computing Accurate Probabilistic Estimates of One-D Entropy from Equiprobable Random Samples.” *Entropy*, vol. 23, no. 6, June 2021, p. 740, doi:10.3390/e23060740.

Hammond, G. E., et al. “Evaluating the Performance of Parallel Subsurface Simulators: An Illustrative Example with PFLOTRAN.” *Wiley Online Library*, vol. 50, no. 1, Jan. 2014, pp. 208–28, doi:10.1002/2012WR013483.

Hare, Danielle K., et al. “A Comparison of Thermal Infrared to Fiber-Optic Distributed Temperature Sensing for Evaluation of Groundwater Discharge to Surface Water.” *Journal of Hydrology*, vol. 530, 2015, pp. 153–66, doi:10.1016/j.jhydrol.2015.09.059.

Hatch, Christine E., et al. “Quantifying Surface Water–Groundwater Interactions Using Time Series Analysis of Streambed Thermal Records: Method Development.” *Water Resources Research*, vol. 42, no. 10, John Wiley & Sons, Ltd, Oct. 2006, p. 10410, doi:10.1029/2005WR004787.

Healy, Richard W., and Anne D. Ronan. “Documentation of Computer Program VS2DH for Simulation of Energy Transport in Variably Saturated Porous Media.” *US Geol Surv Water-Resour Invest Rep 96-4230*, 1996, p. 36, doi:Cited By (since 1996) 24\rExport Date 4 April

2012.

Hochreiter, Sepp, and J. ¨. Urgen Schmidhuber. "Long Short-Term Memory." *MIT Press*, https://www.mitpressjournals.org/doi/abs/10.1162/neco.1997.9.8.1735. Accessed 11 Mar. 2020.

Javadi, S., et al. "Classification of Aquifer Vulnerability Using K-Means Cluster Analysis." *Journal of Hydrology*, vol. 549, Elsevier B.V., June 2017, pp. 27–37, doi:10.1016/j.jhydrol.2017.03.060.

Keery, John, et al. "Temporal and Spatial Variability of Groundwater-Surface Water Fluxes: Development and Application of an Analytical Method Using Temperature Time Series." *Journal of Hydrology*, vol. 336, no. 1–2, 2007, pp. 1–16, doi:10.1016/j.jhydrol.2006.12.003.

Kikuchi, C. P., and T. P. A. Ferré. "Analysis of Subsurface Temperature Data to Quantify Groundwater Recharge Rates in a Closed Altiplano Basin, Northern Chile." *Hydrogeology Journal*, vol. 25, no. 1, 2017, pp. 103–21, doi:10.1007/s10040-016-1472-1.

Kingma, Diederik P., and Jimmy Lei Ba. "Adam: A Method for Stochastic Optimization." *3rd International Conference on Learning Representations, ICLR 2015 - Conference Track Proceedings*, International Conference on Learning Representations, ICLR, 2015.

Kratzert, Frederik, et al. "Toward Improved Predictions in Ungauged Basins: Exploiting the Power of Machine Learning." *Water Resources Research*, vol. 55, no. 12, Blackwell Publishing Ltd, Dec. 2019, pp. 11344–54, doi:10.1029/2019WR026065.

Labaky, W., et al. "Field Comparison of the Point Velocity Probe with Other Groundwater Velocity Measurement Methods." *Water Resources Research*, vol. 45, no. 4, John Wiley & Sons, Ltd, Apr. 2009, doi:10.1029/2008WR007066.

Lai, Vivian, et al. *Many Faces of Feature Importance: Comparing Built-in and Post-Hoc Feature Importance in Text Classification*. Association for Computational Linguistics (ACL), 2019, pp. 486–95, doi:10.18653/v1/d19-1046.

Lautz, Laura K. "Observing Temporal Patterns of Vertical Flux through Streambed Sediments Using Time-Series Analysis of Temperature Records." *Journal of Hydrology*, vol. 464–465, 2012, pp. 199–215, doi:10.1016/j.jhydrol.2012.07.006.

Lewandowski, Jörg, et al. "A Heat Pulse Technique for the Determination of Small-Scale Flow Directions and Flow Velocities in the Streambed of Sand-Bed Streams." *Hydrological Processes*, vol. 25, no. 20, Sept. 2011, pp. 3244–55, doi:10.1002/hyp.8062.

Lichtner, Peter C., et al. *PFLOTRAN User Manual: A Massively Parallel Reactive Flow and Transport Model for Describing Surface and Subsurface Processes*. 2015, doi:10.2172/1168703.

López-Acosta, NP. "Numerical and Analytical Methods for the Analysis of Flow of Water Through Soils and Earth Structures." *Contaminant and Resources Managment*, 2016, p. 91, https://books.google.com/books?hl=en&lr=&id=63uQDwAAQBAJ&oi=fnd&pg=PA91&dq=López-Acosta,+Norma+Patricia.+%22Numerical+and+Analytical+Methods+for+the+Analysis+of+Flow+of+Water+Through+Soils+and+Earth+Structures.%22+Groundwater:+Contaminant+and+Resource+Man.

Luo, Yixin, and Fan Yang. "Deep Learning with Noise." *Pdfs.Semanticscholar.Org*, 2015, pp. 1–9, https://pdfs.semanticscholar.org/d79b/a428e1cf1b8aa5d320a93166315bb30b4765.pdf.

Mamer, Ethan A., and Christopher S. Lowry. "Locating and Quantifying Spatially Distributed Groundwater/Surface Water Interactions Using Temperature Signals with Paired Fiber-Optic Cables." *Water Resources Research*, vol. 49, no. 11, Nov. 2013, pp. 7670–80, doi:10.1002/2013WR014235.

Meyer, C. .. *Thermodynamics and Transport Properties of Steam: Comprising Tables and Charts for Steam and Water*. American Society of Mechanical Engineers, 1968.

Moghaddam, MA, et al. *Applying Simple Machine Learning Tools to Infer Streambed Flux from Subsurface Pressure and Temperature Observations*. Earth and Space Science Open Archive, 2020, doi:10.1002/ESSOAR.10502715.1.

Moghaddam, Mohammad A., et al. "Can Deep Learning Extract Useful Information about Energy Dissipation and Effective Hydraulic Conductivity from Gridded Conductivity Fields?" *Water 2021, Vol. 13, Page 1668*, vol. 13, no. 12, Multidisciplinary Digital Publishing Institute, June 2021, p. 1668, doi:10.3390/W13121668.

Molnar, Christoph. "Interpretable Machine Learning." *Lulu. Com*, 2019, https://books.google.com/books?hl=en&lr=&id=jBm3DwAAQBAJ&oi=fnd&pg=PP1&dq=Interpretable+Machine+Learning+A+Guide+for+Making+Black+Box+Models+Explainable+Christoph+Molnar&ots=EfzTZrKBP1&sig=T17272tPrXLZ2VJMLy_f4hdhJw0.

---. *Quantifying Model Complexity via Functional Decomposition for Better Post-Hoc Interpretability*. Apr. 2019, http://arxiv.org/abs/1904.03867.

Montgomery, DC. *Design and Analysis of Experiments*. 2017, https://books.google.com/books?hl=en&lr=&id=Py7bDgAAQBAJ&oi=fnd&pg=PA1&dq=Montgomery,+Douglas+C.+Design+and+Analysis+of+Experiments.+9th+ed.,+John+Wiley+%26+Sons,+Inc.,+2017.&ots=X7s1oZJV44&sig=yzV-6w_j3rmJgEK49HcJcBIlxzU.

Munz, Matthias, and Christian Schmidt. "Estimation of Vertical Water Fluxes from Temperature Time Series by the Inverse Numerical Computer Program FLUX-BOT." *Hydrological Processes*, vol. 31, no. 15, John Wiley and Sons Ltd, July 2017, pp. 2713–24, doi:10.1002/hyp.11198.

Najafabadi, Maryam M., et al. "Deep Learning Applications and Challenges in Big Data Analytics." *Journal of Big Data*, vol. 2, no. 1, SpringerOpen, Dec. 2015, doi:10.1186/s40537-014-0007-7.

Natekin, Alexey, and Alois Knoll. "Gradient Boosting Machines, a Tutorial." *Frontiers in Neurorobotics*, vol. 7, no. DEC, Frontiers Research Foundation, 2013, doi:10.3389/fnbot.2013.00021.

Nettleton, David F., et al. "A Study of the Effect of Different Types of Noise on the Precision of Supervised Learning Techniques." *Artificial Intelligence Review*, vol. 33, no. 3–4, Kluwer Academic Publishers, 2010, pp. 275–306, doi:10.1007/s10462-010-9156-z.

Nielsen, Didrik. *Tree Boosting With XGBoost Why Does XGBoost Win "Every" Machine Learning Competition?* 2016, https://ntnuopen.ntnu.no/ntnu-xmlui/bitstream/handle/11250/2433761/16128_FULLTEXT.pdf.

Noh, Hyeonwoo, et al. "Regularizing Deep Neural Networks by Noise: Its Interpretation and Optimization." *Advances in Neural Information Processing Systems*, vol. 2017-Decem, 2017, pp. 5110–19, http://papers.nips.cc/paper/7096-regularizing-deep-neural-networks-by-noise-its-interpretation-and-optimization.

Onderka, Milan, et al. "Seepage Velocities Derived from Thermal Records Using Wavelet Analysis." *Journal of Hydrology*, vol. 479, 2013, pp. 64–74, doi:10.1016/j.jhydrol.2012.11.022.

Pelissier, Craig, et al. *Combining Parametric Land Surface Models with Machine Learning*. Feb. 2020, http://arxiv.org/abs/2002.06141.

Pionke, H. B., et al. "Chemical-hydrologic Interactions in the Near-stream Zone." *Water Resources Research*, vol. 24, no. 7, 1988, pp. 1101–10, doi:10.1029/WR024i007p01101.

Rau, Gabriel C., et al. "Assessing the Accuracy of 1-D Analytical Heat Tracing for Estimating near-Surface Sediment Thermal Diffusivity and Water Flux under Transient Conditions." *Journal of Geophysical Research: Earth Surface*, vol. 120, no. 8, Blackwell Publishing Ltd, Aug. 2015, pp. 1551–73, doi:10.1002/2015JF003466.

Ribeiro, Marco Tulio, et al. "'Why Should i Trust You?' Explaining the Predictions of Any Classifier." *Proceedings of the ACM SIGKDD International Conference on Knowledge Discovery and Data Mining*, vol. 13-17-Augu, Association for Computing Machinery, 2016, pp. 1135–44, doi:10.1145/2939672.2939778.

Rohmat, Faizal I. W., et al. "Deep Learning for Compute-Efficient Modeling of BMP Impacts on Stream- Aquifer Exchange and Water Law Compliance in an Irrigated River Basin." *Environmental Modelling and Software*, vol. 122, 2019, doi:10.1016/j.envsoft.2019.104529.

Rosenberry, D. O., et al. "Use of Monitoring Wells , Portable Piezometers , and Seepage Meters to Quantify Flow Between Surface Water and Ground Water." *Field Techniques for Estimating Water Fluxes Between Surface Water and Ground Water*, 2008, p. 128.

Selvin, Sreelekshmy, et al. "Stock Price Prediction Using LSTM, RNN and CNN-Sliding Window Model." *2017 International Conference on Advances in Computing, Communications and Informatics, ICACCI 2017*, vol. 2017-Janua, 2017, pp. 1643–47, doi:10.1109/ICACCI.2017.8126078.

Shen, Chaopeng. "A Transdisciplinary Review of Deep Learning Research and Its Relevance for Water Resources Scientists." *Water Resources Research*, vol. 54, no. 11, Blackwell Publishing Ltd, 1 Nov. 2018, pp. 8558–93, doi:10.1029/2018WR022643.

Song, Yang, et al. "Assessment of Snowfall Accumulation from Satellite and Reanalysis Products Using SNOTEL Observations in Alaska." *Remote Sensing*, vol. 13, no. 15, July 2021, p. 2922, doi:10.3390/rs13152922.

Sophocleous, Marios. "Interactions between Groundwater and Surface Water: The State of the Science." *Hydrogeology Journal*, vol. 10, no. 1, Feb. 2002, pp. 52–67, doi:10.1007/s10040-001-0170-8.

Soto-López, Carlos D., et al. "Effects of Measurement Resolution on the Analysis of Temperature Time Series for Stream-Aquifer Flux Estimation." *Water Resources Research*, vol. 47, no. 12, John Wiley & Sons, Ltd, Dec. 2011, doi:10.1029/2011WR010834.

Tang, Xianzhe, et al. “A Spatial Assessment of Urban Waterlogging Risk Based on a Weighted Naïve Bayes Classifier.” *Science of the Total Environment*, vol. 630, Elsevier B.V., July 2018, pp. 264–74, doi:10.1016/j.scitotenv.2018.02.172.

Therrien, R., et al. *HydroSphere A Three-Dimensional Numerical Model Describing Fully-Integrated Subsurface and Overland Flow and Solute Transport*. 2003.

Triana, Enrique, et al. “River GeoDSS for Agroenvironmental Enhancement of Colorado’s Lower Arkansas River Basin. I: Model Development and Calibration.” *Journal of Water Resources Planning and Management*, vol. 136, no. 2, Mar. 2010, pp. 177–89, doi:10.1061/(ASCE)WR.1943-5452.0000025.

Triska, Frank J., et al. “The Role of Water Exchange between a Stream Channel and Its Hyporheic Zone in Nitrogen Cycling at the Terrestrial—Aquatic Interface.” *Nutrient Dynamics and Retention in Land/Water Ecotones of Lowland, Temperate Lakes and Rivers*, Springer Netherlands, 1993, pp. 167–84, doi:10.1007/978-94-011-1602-2_20.

Vandersteen, G., et al. “Determining Groundwater-Surface Water Exchange from Temperature-Time Series: Combining a Local Polynomial Method with a Maximum Likelihood Estimator.” *Water Resources Research*, vol. 51, no. 2, Blackwell Publishing Ltd, 2015, pp. 922–39, doi:10.1002/2014WR015994.

Vezhnevets, Alexander, and Olga Barinova. “Avoiding Boosting Overfitting by Removing Confusing Samples.” *Lecture Notes in Computer Science (Including Subseries Lecture Notes in Artificial Intelligence and Lecture Notes in Bioinformatics)*, vol. 4701 LNAI, 2007, pp. 430–41, doi:10.1007/978-3-540-74958-5_40.

Voytek, Emily B., et al. “1DTempPro: Analyzing Temperature Profiles for Groundwater/Surface-Water Exchange.” *Groundwater*, vol. 52, no. 2, Blackwell Publishing Ltd, 2014, pp. 298–302, doi:10.1111/gwat.12051.

Xu, Hongshi, et al. “Urban Flooding Risk Assessment Based on an Integrated K-Means Cluster Algorithm and Improved Entropy Weight Method in the Region of Haikou, China.” *Journal of Hydrology*, vol. 563, Elsevier B.V., Aug. 2018, pp. 975–86, doi:10.1016/j.jhydrol.2018.06.060.

Zhao, Bendong, et al. “Convolutional Neural Networks for Time Series Classification.” *Journal of Systems Engineering and Electronics*, vol. 28, no. 1, Beijing Institute of Aerospace Information, Feb. 2017, pp. 162–69, doi:10.21629/JSEE.2017.01.18.

Zheng, Huiting, et al. “Short-Term Load Forecasting Using EMD-LSTM Neural Networks with a Xgboost Algorithm for Feature Importance Evaluation.” *Energies*, vol. 10, no. 8, MDPI AG, Aug. 2017, p. 1168, doi:10.3390/en10081168.

Zolfaghari, Reza, et al. “A Comparison between Different Windows in Spectral and Cross Spectral Analysis Techniques with Kalman Filtering for Estimating Power Quality Indices.” *Electric Power Systems Research*, vol. 84, no. 1, Elsevier, Mar. 2012, pp. 128–34, doi:10.1016/j.epsr.2011.10.017.